\documentclass[a4paper,fleqn]{cas-dc}

\usepackage[authoryear]{natbib}

\def\tsc#1{\csdef{#1}{\textsc{\lowercase{#1}}\xspace}}
\tsc{WGM}
\tsc{QE}
\tsc{EP}
\tsc{PMS}
\tsc{BEC}
\tsc{DE}

\usepackage{multirow}
\usepackage{wrapfig}

\usepackage{CJKutf8}

\usepackage{graphics}
\usepackage{graphicx}
\usepackage{subfig}
\usepackage{enumitem}
\usepackage{balance}
\usepackage{arydshln}
\usepackage{amsmath}
\definecolor{zptu}{RGB}{18, 141, 21}

\definecolor{wx}{RGB}{18, 141, 21}

\begin{document}
\let\WriteBookmarks\relax
\def\floatpagepagefraction{1}
\def\textpagefraction{.001}

\shorttitle{On the Shortcut Learning in Multilingual Neural Machine Translation}

\shortauthors{Wang et~al.}

\title [mode = title]{On the Shortcut Learning in Multilingual Neural Machine Translation}                      




%
\author[1]{Wenxuan Wang}[orcid=0000-0002-9803-8204]


\ead{wxwang@cse.cuhk.edu.hk}


\credit{Conceptualization of this study, Methodology, Software}

\affiliation[1]{organization={The Chinese University of Hong Kong},
    city={Hong Kong},
    country={China}}

\author[2]{Wenxiang Jiao}
\ead{wenxiangjiaonju@gmail.com}

\author[1]{Jen-tse Huang}
\ead{jthuang@cse.cuhk.edu.hk}

\credit{Data curation, Writing - Original draft preparation}

\affiliation[2]{organization={Tencent AI Lab},
    city={Shenzhen},
    state={Guangdong},
    country={China}}

\author%
[2]
{Zhaopeng Tu}
\ead{tuzhaopeng@gmail.com}

\cormark[1]

\author%
[1]
{Michael Lyu}
\ead{lyu@cse.cuhk.edu.hk}

\cortext[cor1]{Corresponding author}



\begin{abstract}
In this study, we revisit the commonly-cited off-target issue in multilingual neural machine translation (MNMT). By carefully designing experiments on different MNMT scenarios and models, we attribute the off-target issue to the overfitting of the shortcuts of (non-centric, centric) language mappings. Specifically, the learned shortcuts biases MNMT to mistakenly translate non-centric languages into the centric language instead of the expected non-centric language for zero-shot translation.
Analyses on learning dynamics show that the shortcut learning generally occurs in the later stage of model training, and multilingual pretraining accelerates and aggravates the shortcut learning. 
Based on these observations, we propose a simple and effective training strategy to eliminate the shortcuts in MNMT models by leveraging the forgetting nature of model training.
The only difference from the standard training is that we remove the training instances that may induce the shortcut learning in the later stage of model training.
Without introducing any additional data and computational costs, our approach can consistently and significantly improve the zero-shot translation performance by alleviating the shortcut learning for different MNMT models and benchmarks.
\end{abstract}



\begin{keywords}
Multilingual Translation \sep Zero-Shot Translation \sep Off-Target Issue \sep Pre-Training 
\end{keywords}

\maketitle

\section{Introduction}

Multilingual neural machine translation~(MNMT) aims to translate between any two languages with a unified model~\citep{Johnson:2017:TACL,Aharoni:2019:NAACL}. It is appealing due to its efficient deployment and effective cross-lingual knowledge transfer, which enables translations between unseen language pairs, i.e., zero-shot translation.
Zero-shot translation is an important capability of MNMT models since it covers most of the possible translation directions, which are difficult and expensive to be covered by the human-annotated training data.
However, previous studies demonstrate that zero-shot translation often suffers from the off-target issue~\citep{Ha:2016:arXiv,Gu:2019:ACL,Zhang2020ACL}, where the MNMT model tends to translate into other languages rather than the expected target language.

In this paper, we revisit the off-target issues in MNMT with a single centric language, {where parallel training data is only available for directions between the centric language and other languages. This setting has been commonly adopted in research~\citep{Johnson:2017:TACL,Gu:2019:ACL,Zhang2020ACL,Tang:2021:ACL,Wenzek2021WMT} 
and commercial scenes (e.g., business export to overseas).
In such scenarios, zero-shot translation aims to translate between non-centric languages.
We vary the centric language of training dataset, and find that the off-target issues for zero-shot translation are mainly in the corresponding centric language in all cases.
{In addition, although multilingual pretraining has been widely adopted to improve the supervised translation performance~\citep{Liu2020mbart, Tang:2021:ACL}, we find that}
it sacrifices the zero-shot translation performance by introducing remarkably more off-target issues. This finding does not conform to previous studies~\citep{Brown2020LanguageMA,Conneau2020UnsupervisedCR}, which have shown that pretraining improves the generalization ability in zero-shot scenarios.

To better understand these observations, we analyze the learning dynamics and find that the performance of zero-shot translation is fluctuating during training. While MNMT models keep improving cross-lingual transformation ability for zero-shot translation, they ignore the zero-shot language mapping in model training by overfitting to the shortcut of supervised (non-centric, centric) language mapping, which mainly occurs at the late stage of training.
Multilingual pretraining accelerates and aggravates the shortcut learning by introducing another type of shortcut (i.e., the copy of source language) due to the denoising auto-encoding objective~\citep{Liu2020mbart}. The commonality between the two types of shortcuts, i.e., both ignore the target language tag when the source languages are non-centric, enables a fast transformation from the copy pattern embedded in the pretraining initialization to the (non-centric, centric) mapping pattern embedded in the MNMT data during finetuning.
Accordingly, the off-target issue becomes more severe for MNMT models with pretraining (e.g., 94.6\% of zero-shot translations are off-target).




Based on these understandings, we propose a simple and effective training strategy, named {\em generalization training}, to break the shortcut data patterns.
The starting point of our approach is an observation: NMT models tend to gradually forget previously learned knowledge and swing to fit the new training examples during training~\citep{Shao:2022:ACL}.
We leverage the forgetting nature of model training to forget the overfitted (non-centric, centric) language mapping.
Specifically, we divide training process of MNMT into two phases: (1) standard training phase (the first $N-G$ training steps) to train the model on the full training data; and (2) {\em generalization training phase} (the last $G$ training steps) to train the model only on the training examples of (centric, non-centric) language pair. The shortcuts of (non-centric, centric) language mapping no longer exist in the later stage of training, thus would be forgotten by the MNMT model.
Our approach does not introduce any additional computational cost and code modification, which makes it accommodate existing MNMT models seamlessly.

We conduct comprehensive experiments on several benchmarks that vary in language distribution (balanced and imbalanced) and language number (e.g., 6, 16, and 50).
Experimental results show that our approach can consistently and significantly improve zero-shot translation performance, and maintain the performance of supervised translation. We also compare with related work on improving zero-shot translation~\citep{Liu2021ImprovingZT,Wu2021LanguageTM}: our approach outperforms both strong baselines, and combining them together can further improve model performance.
Further analysis dispels the possibility of gradually reducing the number of (non-centric, centric) instances during training: using as few as 0.1\% of such instances hardly mitigate the off-target issues for MNMT with pretraining.

\paragraph{Contributions} Our main contributions are:
\begin{itemize}
\item We link the off-target issue in zero-shot translation to the shortcut learning of MNMT on the supervised language mapping.
\item We find that multilingual pretraining accelerates and aggravates the shortcut learning, which leads to worse generalization performance on zero-shot translation.
\item We propose a simple and effective training approach to improve the generalization ability on zero-shot translation.
\end{itemize}

\section{Related Work}
\label{sec:related_works}

Recent studies have shown that many deep learning problems can be seen as different symptoms of the same underlying problem: {\em shortcut learning}~\citep{geirhos2020shortcut}. Shortcuts are decision rules that perform well on standard benchmarks but fail to transfer to more challenging testing conditions.
Our study connects the commonly-cited off-target issues in MNMT to shortcut learning: the shortcut of supervised language mapping fails to transfer to zero-shot translation.

Shortcut learning has been explored in many NLP tasks including language inference \citep{shortcut-nli1}, reading comprehension \citep{shortcut-rc}, question answering \citep{shortcut-qa}, evaluation of text generation \citep{shortcut-qe}.
Shortcuts learned by the models usually take the form of learning the superficial correlation between simple statistics and the label, which is also known as non-robust features.
For example, 
in reading comprehension task, models mainly focus on the lexical matching of words between the question and the original passage \citep{shortcut-rc}.
Since shortcut learning overfits to the artifacts of the training data, it will hurt model performance when fed with out-of-distribution data and hurt the robustness against adversarial attacks~\citep{du2022shortcut}.

Shortcut learning has rarely been studied in multilingual neural machine translation. The most relevant work is by \cite{Gu:2019:ACL}, which attributed the poor performance of zero-shot translation to the spurious correlation in data. Our work differs from theirs in several aspects:
(1) They only showed that zero-shot translation tends to translate into ``wrong target languages'', while we refine the ``wrong target languages'' to the centric languages, which allows us to locate the underlying reasons (i.e., the commonly-used single centric language setting). 
(2) We find that multilingual pretraining harms the performance of zero-shot translation, which has not been revealed in their study.
(3) They adopted back-translation and decoder pretraining to regularize the spurious correlation, which require additional computation costs for data augmentation and model training. In contrast, our generalization training is more efficient by only making a slight change to the standard training.


\section{Preliminary}

\subsection{Multilingual Machine Translation}
\label{sec:mnmt}

Neural Machine Translation is a representative task in Artificial Intelligence that uses neuron networks to conduct translation~\citep{Sutskever2014SequenceTS, Bahdanau2014NeuralMT, Mi2022ImprovingDA, IranzoSnchez2021StreamingCS, Zhang2023FewshotLP, Yu2021AnIM}. Multilingual neural machine translation~ aims to translate between any two languages with a unified model~\citep{Johnson:2017:TACL,Aharoni:2019:NAACL}. Specifically, an MNMT model is trained on a dataset consisting of parallel sentences in multiple language pairs. 
Given a source sentence $\mathbf{x}^s$ in language $s$ and its translation $\mathbf{y}^t$ in language $t$, the MNMT model translates as below:
\begin{align*}
    \mathbf{H}_{enc} &= \mathrm{Encoder}([\mathbf{x}^s]), \\
    \mathbf{H}_{dec} &= \mathrm{Decoder}([\mathbf{y}^t], \mathbf{H}_{enc}).
\end{align*}
The model is trained with maximum likelihood estimation on the multilingual datasets:
\begin{align*}
    \mathcal{L} =-\sum\nolimits_{i}^{N}\sum\nolimits_{(\mathbf{x},\mathbf{y})\in D_i} \log P([\mathbf{y}^t]|\mathbf{H}_{dec}(\mathbf{x},\mathbf{y})),
\end{align*}
where $N$ is the number of language pairs and $D_i$ is the training instances in the $i$-th language pair.

\paragraph{Zero-Shot Translation.} 
One appealing capability of MNMT is translation between language pairs that do not exist in the training data, i.e., zero-shot translation. {For example, an MNMT model trained on German-English and English-French data is able to
translate from German to French.} However, the performance of zero-shot translation generally lags behind the supervised translation due to the lack of explicit signal during training. 
Improving zero-shot translation is critical for MNMT, and has received a lot of attention in recent years~\citep{Gu:2019:ACL,Zhang2020ACL,Wang:2021:EMNLP}.

\paragraph{Functionalities of MNMT.}
Comparing with the bilingual NMT that only models {\em cross-lingual transformation}, MNMT needs to learn additional functionality of mapping from the source language to the target language (i.e., {\bf \em language mapping} ). 

\begin{figure}[h]
    \centering
    \subfloat[Functionalities of Bilingual Translation]{
    \includegraphics[width=0.9\columnwidth]{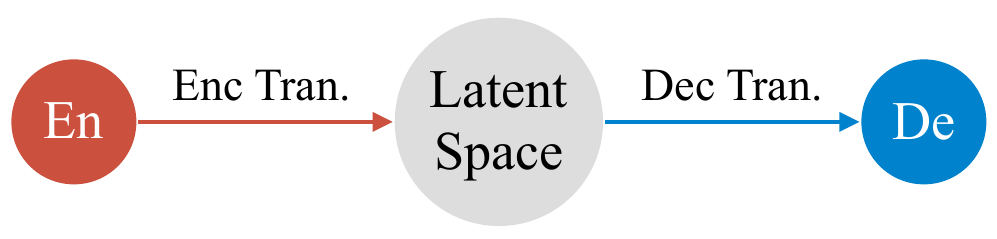}
    }\\
    \subfloat[Functionalities of Multilingual Translation]{
    \includegraphics[width=0.9\columnwidth]{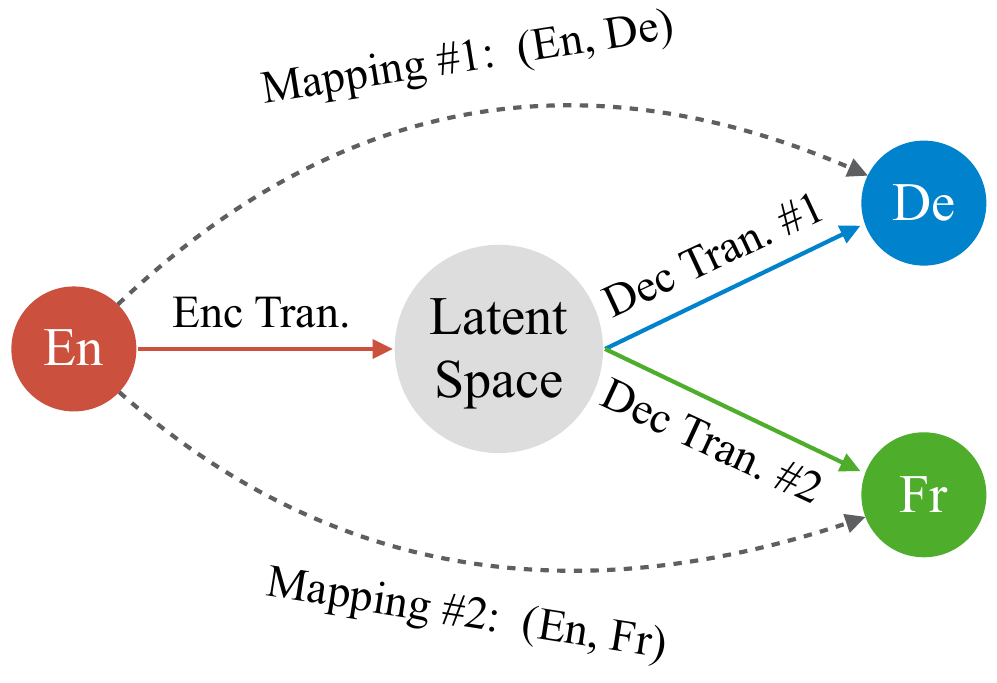}
    }
    \caption{Illustration of cross-lingual transformation and language mapping in multilingual translation.}
    \label{fig:xlt-lm}
\end{figure}

We take Figure~\ref{fig:xlt-lm} as an example to elaborate the concepts of cross-lingual transformation and language mapping in MNMT. In bilingual translation shown as Figure~\ref{fig:xlt-lm}(a), the source language and target language are fixed (e.g. En and De). The main functionality of bilingual translation is cross-lingual transformation, which encodes the source sentence into latent representation space (e.g. ``Enc Tran.'') and decodes from the latent space into the target language (e.g. ``Dec Tran.'').

In multilingual translation (Figure~\ref{fig:xlt-lm}(b)), the source sentence in English can be translated into either German (e.g. ``Dec Tran \#1'') or French (``Dec Tran \#2''). Accordingly, MNMT models needs a strategy to explicitly specify the translation path (i.e. language mapping), which is generally implemented by attaching language tags~(e.g. ``\texttt{\_\_En\_\_}'' for En, and ``\texttt{\_\_De\_\_}'' for De) to the parallel data. The (\texttt{\_\_En\_\_}, \texttt{\_\_De\_\_}) tag pair will guide MNMT models to choose mapping \#1, and the (\texttt{\_\_En\_\_}, \texttt{\_\_Fr\_\_}) tag pair will guide MNMT models to choose mapping \#2.

It is difficult for MNMT to learn the language mapping for zero-shot translation, since the language pair never exists in the training data. Accordingly, previous studies~\citep{Zhang2020ACL,Wang:2021:EMNLP,Yang2021ImprovingMT} have reported that zero-shot translation often suffers from the off-target issues (i.e., translating into wrong target language) on the representative benchmarks. In this work, we revisit this problem and identify a key reason that is responsible for the off-target phenomenon.

\subsection{Multilingual Pretraining}
There has been a wealth of research over the past several years on sequence-to-sequence (Seq2Seq) pretraining models for machine translation, e.g., MASS~\citep{song2019mass}, BART~\citep{lewis2020bart}, and mBART~\citep{Liu2020mbart}.
Generally, Seq2Seq pretraining model (e.g., mBART) shares the same architecture and loss format with standard MNMT models. The main difference is that the source sentence is a corruption of the target sentence in the same language $s$: ${\bf x}^s = g({\bf x}^s)$, where $g$ is a noising function (e.g., randomly masking or reordering tokens).

\paragraph{Discrepancy Between Seq2Seq Pretraining and NMT.}
Seq2Seq pretraining models that are trained on large-scale multilingual language data (i.e., mBART), are generally used to initialize the MNMT models, leading to significant improvement on translation performance across various language pairs.
However, recent studies identified several critical side-effects of Seq2Seq pretraining models due to the objective discrepancy between pretraining and translation, e.g., over-copying issues~\citep{Liu:2021:ACL} and over-estimation issues~\citep{Wang:2022:ACL}.
The pretraining objective learns to reconstruct a few source tokens (i.e., the corrupted tokens) and copy most of them, while the translation objective learns to translate text from source language to target language.
In this work, we identify another side-effect of pretraining model on the off-target issues in multilingual translation.

\subsection{Experimental Setup}

\begin{table}[h!]
\caption{Sizes of (a) Imbalanced CC16 and (b) Noisy Imbalanced OPUS50 English-centric dataset.} 
\centering
\scalebox{0.9}{
\subfloat[Imbalanced CC16-En]{
\begin{tabular}{lr lr lr}
\toprule
{\bf Language} & \bf Size & {\bf Language} & \bf Size & {\bf Language} & \bf Size \\
\cmidrule(lr){1-2} \cmidrule(lr){3-4} \cmidrule(lr){5-6}
Spanish &  2.73 &  Russian &  0.93  & Arabic & 0.33  \\
French & 2.20  & Chinese & 0.50 & Japanese & 0.27  \\
German & 1.66 & Indonesian & 0.47 & Korean &  0.12\\
Portuguese & 1.16 &  Romanian & 0.37 & Hindi & 0.10\\
Italian & 0.97 & Vietnamese & 0.33 & Thai & 0.07 \\
\hline
{\bf Total} & 11.05 \\
\bottomrule
\end{tabular}
}
}
\\
\scalebox{0.9}{
\subfloat[Noisy Imbalanced OPUS50-En]{
\begin{tabular}{lr lr lr}
\toprule
{\bf Language} & \bf Size & {\bf Language} & \bf Size & {\bf Language} & \bf Size\\
\cmidrule(lr){1-2} \cmidrule(lr){3-4} \cmidrule(lr){5-6}
German  & 1.0  &  Estonian  & 1.0 &     Hindi    &   0.5\\ 
French  & 1.0  &  Latvian   & 1.0 & Nepali    &   0.4\\
Chinese & 1.0  &  Macedonian & 1.0 &   Xhosa  &   0.4\\
Romanian & 1.0 &  Persian   & 1.0 &  Georgian    &   0.4\\  
Japanese & 1.0 &  Sinhala   & 1.0 & Azerbaijani    &   0.3\\
Turkish & 1.0 & Ukrainian & 1.0 & Afrikaans   &   0.3 \\
Russian     &  1.0   & Croatian & 1.0 & Gujarati  &   0.3\\
Italian     &  1.0   & Finnish & 1.0 & Tamil   &   0.2\\
Indonesian  &  1.0   & Bengali & 1.0 & Khmer   &   0.1 \\
Spanish     & 1.0 & Lithuanian  & 1.0 & Kazakh  &   0.08\\
Vietnamese & 1.0 & Dutch & 1.0 & Pashto   &   0.08 \\
Arabic & 1.0  & Polish & 1.0 & Telugu    &   0.06 \\
Portuguese  & 1.0  & Slovenian & 1.0 & Marathi   &   0.03 \\
Korean & 1.0 & Czech & 1.0   &  Burmese &   0.02 \\
Thai & 1.0  &  Urdu  &   0.8 & \\
Swedish & 1.0  & Malayalam & 0.8 \\
\cline{5-6}
Hebrew & 1.0  &  Galician &   0.5 & {\bf Total}  &   36.07\\
\bottomrule
\end{tabular}
}
}
\label{tab:dataset}
\end{table}

\paragraph{Training Data.}
The training datasets (see Table~\ref{tab:dataset}  for data statistics) include:
\begin{itemize}
    \item \textbf{Balanced CCMatrix Datasets with Different Centric Languages.} 
    We construct six balanced datasets, where each distinct language from (En, De, Fr, Ro, Ja, and Zh) serves as the single centric language.
    We sample 1.0M sentence pairs from the CCMatrix~\citep{Schwenk2021CCMatrixMB} data for each language pair ({\em Balanced CC6-X}, 5M). 
    \item \textbf{Imbalanced CCmatrix Datasets.} We simulate a common situation in MNMT with imbalanced training data. We randomly sample the subsets from the CCMatrix data to construct an imbalanced English-centric dataset ({\em Imbalanced CC16-En}, 11M) that consists of 16 languages.
    \item \textbf{Noisy Imbalanced OPUS Datasets.} 
    \citep{Zhang2020ACL} propose OPUS-100 dataset that consists of 55M English-centric sentence pairs covering 100 languages. 
    Previous studies~\citep{Wang:2022:arXiv} have revealed that for 5.8\% of the training examples in the OPUS100 data, the target sentences are in the source language.
    We select the 50 languages used in mBART50~\citep{Tang:2021:ACL} to construct an imbalanced dataset ({\em Noisy ImBalanced OPUS50-En}, 36M).
\end{itemize}

\paragraph{Evaluation Data.}
To eliminate the content bias across languages, we evaluate the performance of multilingual translation models on the multi-way Flores valid/test set
~\citep{goyal2021flores},
which contains 997/1012 sentences translated into 101 languages.
We report the results of both BLEU scores~\citep{papineni2002bleu} and off-target ratios (OTR) for both supervised and zero-shot translation. 
For example, the CC6-En dataset contains 10 supervised directions (i.e., En-X and X-En) and 20 zero-shot directions (i.e., X-X).
To calculate the off-target ratio in translation output, we employ the \texttt{langid} library\footnote{\url{https://github.com/saffsd/langid.py}}, the most widely used language identification tool with 93.7\% accuracy on 7 dataset across 97 languages, to detect the language of generated sentences.


\paragraph{Model.}
To support both training from scratch and finetuning from pretrained models, we adopt an MNMT model with the same architecture as the mBART50 model
~\citep{Tang:2021:ACL}, which consists of 12 encoder layers and 12 decoder layers with 1024 dimensions. 
We follow the common practices to attach the source language tag to encoder and the target language tag to decoder~\citep{Tang:2021:ACL,Fan2021BeyondEM}.
We use the vocabulary of mBART that is built for 100 languages, which can enable the scaling of languages. 
On the CC-6 datasets, we train the models with 
65K tokens per batch (4096 tokens $\times$ 16 GPU) for 100K updates.
For the CC16 dataset, we enlarge the batch size to 130K tokens for 200K steps. For the OPUS50 dataset, we further enlarge the batch size to 260K tokens for 300K steps.
The finetuning hyper-parameters are from the official recommendation with dropout of $0.3$, label smoothing of $0.2$, and warm-up of 10K steps.
{
Results on smaller model architectures can be found in Appendix~\ref{app:smaller-arch}.
}

\section{Observing Shortcut Learning}

In this section, we establish that the commonly-used datasets with a single centric language is questionable when used for conducting zero-shot translation. We first revisit the off-target issues on the single-centric datasets (\S~\ref{sec:off-target}), and connect them to the shortcut learning on the supervised (non-centric, centric) language mapping (\S~\ref{sec:shortcut}).
We finally empirically analyze the reasons behind the shortcut learning in model training (\S~\ref{sec:model-training}). 


\subsection{Revisiting Off-Target Issues}
\label{sec:off-target}

We revisit the off-target issue from two angles by: (1) varying the centric languages of multilingual translation datasets; and (2) training MNMT models from scratch or finetuning from the mBART50 pretraining model, to offer a more comprehensive understanding, as listed in Table~\ref{tab:centric-langs-one}. 


\begin{table}
\caption{Translation performance (BLEU$\uparrow$) and off-target ratios (OTR$\downarrow$) on 
{\bf balanced CC6 datasets} with single centric language.
``OTR$_C$'' denotes that {off-target ratio on the centric language(s)}.} 
\centering
\begin{tabular}{cc rr rr}
\toprule
\bf Cen. & {\bf Pre-} & \multicolumn{1}{c}{\bf Sup.}  & \multicolumn{3}{c}{\bf Zero-Shot}\\
\cmidrule(lr){3-3}\cmidrule(lr){4-6}
\bf Lang.   & \bf Train & \it BLEU   &   \it BLEU  & \it OTR & \it OTR$_C$\\
\cmidrule(lr){1-2}\cmidrule(lr){3-6}
\multirow{2}{*}{\bf En}  & \texttimes & 37.4 & 15.5 & 36.2 & 35.8 \\
                         & \checkmark & 38.1 & 2.9 & 94.8 & 94.6    \\
\hdashline
\multirow{2}{*}{\bf De}  & \texttimes  & 30.0 & 24.7 & 8.1 & 7.7 \\
                         & \checkmark  & 30.6 & 2.7 & 95.4 & 95.3  \\
\hdashline
\multirow{2}{*}{\bf Fr}  & \texttimes  & 33.5 & 24.2 & 6.8 & 6.4\\
                         & \checkmark  & 34.2 & 10.5 & 33.6 & 33.2 \\
\hdashline
\multirow{2}{*}{\bf Ro}  & \texttimes  & 30.7 & 19.8 & 23.6 & 23.3 \\
                         & \checkmark  & 31.1 & 9.0 &  72.6 & 72.3\\
\hdashline
\multirow{2}{*}{\bf Ja}  & \texttimes  & 28.8 & 19.0  & 23.5 & 23.1 \\
                         & \checkmark  & 29.8 & 14.7 & 49.7 & 49.4 \\
\hdashline
\multirow{2}{*}{\bf Zh}  & \texttimes  & 28.7 & 26.1 & 4.3 & 4.1 \\
                         & \checkmark  & 29.4 & 21.7 & 21.3 & 21.0 \\
\bottomrule
\end{tabular}
\label{tab:centric-langs-one}
\vspace{-10pt}
\end{table}

\paragraph{Off-target translations are mainly on the centric language.} 
While the off-target ratio~(OTR) varies across different datasets, we find that almost all the off-target translations are directed to the centric languages.
Our study connects the off-target issue to the centric language of datasets, which has not been revealed in previous studies.


\paragraph{Pretraining aggravates the off-target issues.}
Pretraining consistently improves the performance of supervised translation, while harms that of zero-shot translation by introducing more off-target issues. 
For example, pretraining produces more than 90\% off-target translations for zero-shot translation on the English- and German-centric datasets.
These results indicate that pretraining harms the generalization ability on zero-shot translation, which will be discussed in the following sections.

\subsection{Shortcut Learning of Language Mapping}
\label{sec:shortcut}

\begin{table}
\caption{Averaged distributions of output languages with different target tags. ``None'' denotes omitting target tag. The source sentences are in French (``Fr'') from the Flores Valid Set.}
\setlength{\tabcolsep}{4pt}
\centering
\scalebox{0.9}{
\begin{tabular}{c lll}
\toprule
\bf Target  &   {\bf Pretrain}  &   \multicolumn{2}{c}{\bf MNMT Model}\\
\cmidrule(lr){3-4}
\bf Tag     &   \bf Model   & {\bf w/o Pretrain} & {\bf w/ Pretrain} \\
\midrule
None    & Fr(100\%) & En(100\%) & En(100\%)\\
\hdashline
Fr      & Fr(100\%) & Fr(97\%), En(3\%)   & En(100\%)\\
De      & Fr(100\%) & De(60\%), En(40\%)  & En(100\%)\\
\hdashline
En      & Fr(100\%) & En(100\%) & En(100\%)\\
\bottomrule
\end{tabular}
}
\label{tab:target-tag-manuipulate-baselines}
\vspace{-10pt}
\end{table}

\begin{figure*}[t!]
    \vspace{-10pt}
    \centering 
    \subfloat[BLEU: w/o Pretrain]{
    \includegraphics[height=0.24\textwidth]{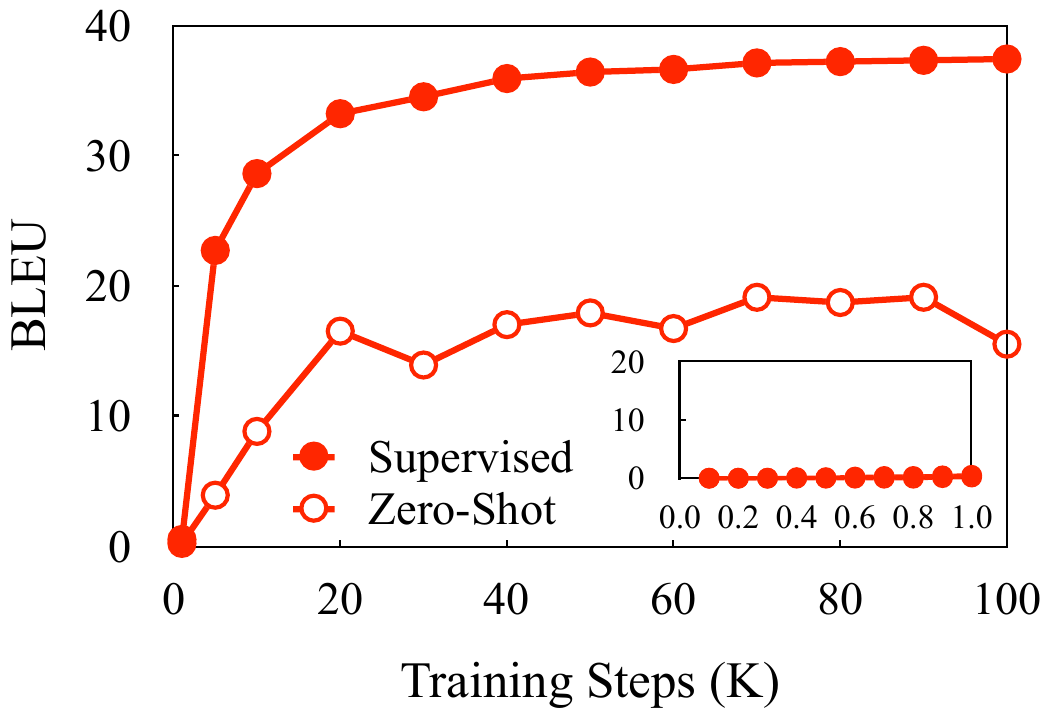}}
    \hspace{0.07\textwidth}
    \subfloat[OTR: w/o Pretrain]{
    \includegraphics[height=0.24\textwidth]{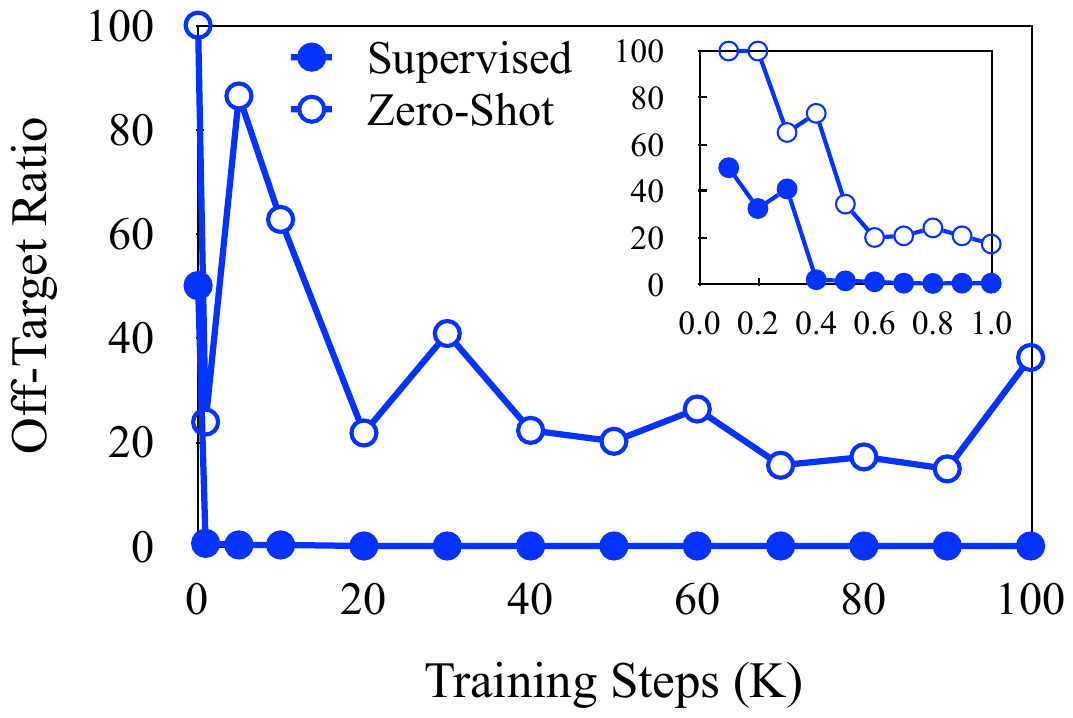}}
    \\
    \subfloat[BLEU: w/ Pretrain]{
    \includegraphics[height=0.24\textwidth]{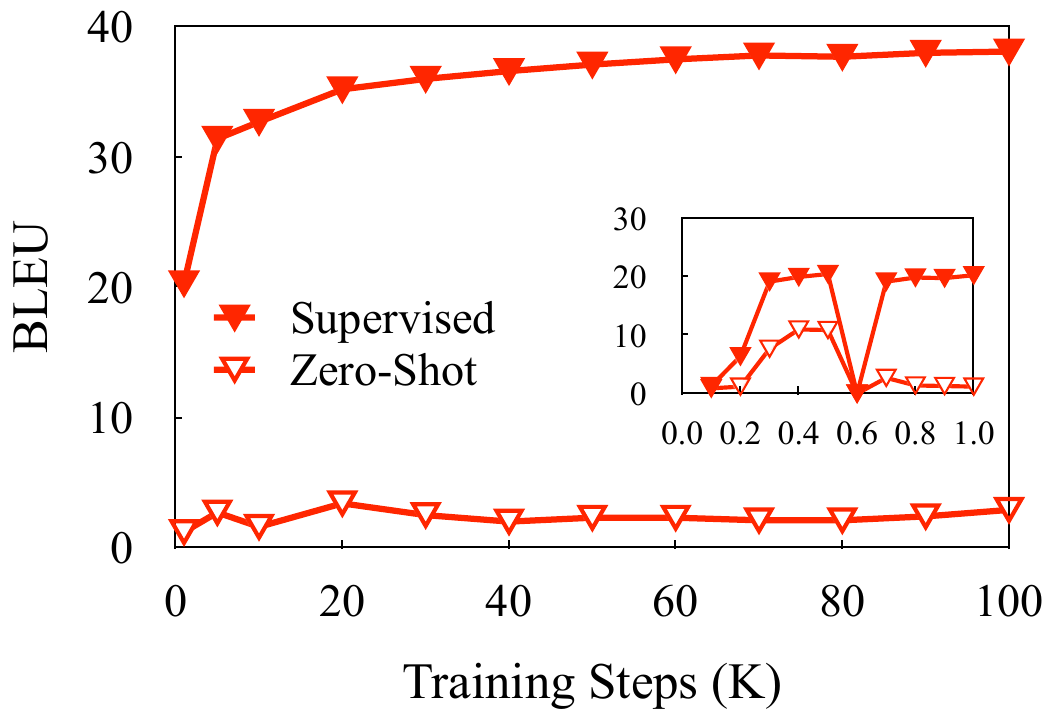}}
    \hspace{0.07\textwidth}
    \subfloat[OTR: w/ Pretrain]{
    \includegraphics[height=0.24\textwidth]{figures/vanilla-learning-otr-no-pretrain-cc6-en-recut.pdf}}
    \caption{Learning curves of the vanilla MNMT models (a,b) w/o pretraining and (c,d) w/ pretraining.} 
    \label{fig:learning-curves-vanilla}
\end{figure*}

Recent work has shown that deep learning models in NLP are highly sensitive to low-level correlations between simple features and specific output labels, leading to over-fitting and lacking of generalization~\citep{Schwartz:2022:NAACL}.
Starting from the finding, we conjecture that {\em MNMT model overfits the supervised language mapping, and lacks generalization of zero-shot language mapping}.
During training, all non-centric languages are translated into the centric language, which may allow the model to overfit the {\bf shortcut} of (non-centric, centric) language mapping.

To validate that MNMT model overfits the shortcut of (non-centric, centric) language mapping, we manipulate the target language tags and identify the language of the generated texts.
Table~\ref{tab:target-tag-manuipulate-baselines} lists the averaged distributions of output languages for translating non-centric languages with different target language tags. 
Given input sentences in French, the pretraining model outputs French sentences regardless of the given target tags (e.g., ``Fr (100\%)''), indicating that pretraining model suffers from more severe shortcut learning problem. This is intuitive, since the shortcut in pretraining -- copy of the source language, is easier to learn. Comparing with the vanilla MNMT model (i.e., ``w/o Pretrain''), the pretrained MNMT model (``w/ Pretrain'') translates all non-centric French sentences into the centric language English for all target tags, showing that pretraining initialization aggravates the shortcut learning in MNMT.
The malfunction of target language tag confirms our research hypothesis on the connection between off-target issues and shortcut of supervised language mapping.



\subsection{Shortcut Learning in Model Training}
\label{sec:model-training}

The above results imply that MNMT models tend to ignore the given target language tag for zero-shot translation in inference. 
In this section, we analyze the training process of MNMT models and show how the model overfits to the shortcut of (non-centric, centric) language mapping.
Unless otherwise stated, all results are reported on the Flores Validation Set for the CC6-En data. The results on CC6-Ro data can be found in Figure~\ref{fig:learning-curves-vanilla-ro} in Appendix, where all conclusions still hold.


\paragraph{The shortcut learning on (non-centric, centric) language mapping occurs at the late training stage (Figures~\ref{fig:learning-curves-vanilla}(a,b)).}
While the supervised translation performance of vanilla MNMT model keeps growing during training, the zero-shot translation performance fluctuates after 20K steps.
The OTR of supervised translation declines to almost 0 at the very beginning of the training (e.g., 0.4K step) and maintains stably in the following training process.
In contrast, the OTR of zero-shot translation first decreases at the early training stage, and reaches 20.1 OTR at the 0.6K step, which is even lower than the finally trained MNMT model (e.g., 35.8 OTR).
Then the OTR of zero-shot translation suddenly increases and fluctuates after 20K steps, showing that the model is biased to translate the non-centric languages into the centric language.

\paragraph{Pretraining {\em accelerates} and {\em aggravates} the shortcut learning 
(Figures~\ref{fig:learning-curves-vanilla}(c,d)).}
Pretraining improves the training of supervised translation with better initialization, at the cost of sacrificing the performance of zero-shot translation (e.g., around 2.4 BLEU).
As shown in the internal small images, pretraining accelerates the shortcut learning: the fluctuation of OTR happens as early as in the first 1K steps. Afterwards, the OTR stays high during the whole training process, which is much more severe than the vanilla model without pretraining.
One interesting finding is that the pretrained MNMT model crashes at the 0.6K step ({\bf inflection point}), where the BLEU scores of both supervised and zero-shot translations decline to 0 and their OTRs reaches 100.

\begin{figure}
    \vspace{-10pt}
    \centering 
    \subfloat[w/o Pretrain]{
    \includegraphics[height=0.24\textwidth]{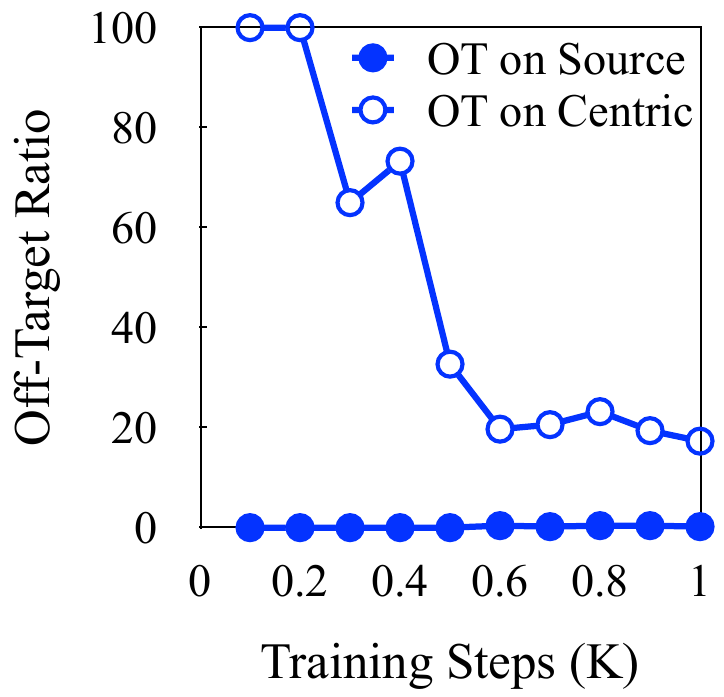}}
    \hfill
    \subfloat[w/ Pretrain]{
    \includegraphics[height=0.24\textwidth]{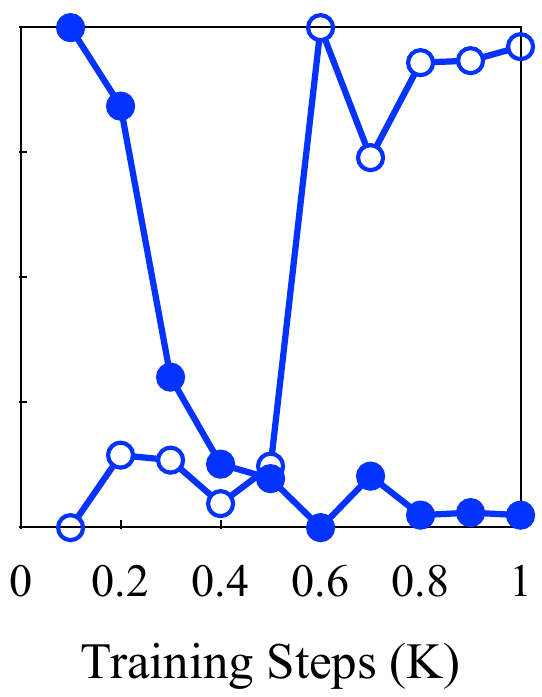}}
    \caption{Ratios of off-target translations on the source language and centric language.} 
    \label{fig:otr-source-centric}
\end{figure}

One possible reason to the inflection point is the transition of shortcuts of language mapping between pretraining and MNMT.
As listed in Table~\ref{tab:target-tag-manuipulate-baselines}, the shortcut of pretraining is copy of source language (e.g., (Fr, Fr)), while that of MNMT is (non-centric, centric) (e.g., (Fr, En)).
A common consequence is that both pretraining and MNMT ignore the target languages when the source languages are non-centric.
When fine-tuning on the MNMT data, this commonality enables a fast transformation from the copy pattern embedded in the pretraining initialization to the (non-centric, centric) mapping pattern embedded in the MNMT data.
Figure~\ref{fig:otr-source-centric} plots the OTRs on source language and centric language for zero-shot translation.
The off-target translations in the vanilla MNMT model (``w/o Pretrain'') are mainly on the centric language. In contrast, the off-target translations in the pretrained MNMT model (``w/ Pretrain'') are mainly on the source language at the beginning, which declines to 0 until the inflection point (i.e., 0.6K step). Afterwards, the off-target translations of the finetuned MNMT model is mainly on the centric language.
Table~\ref{tab:case-vanilla} in Appendix shows translations at different steps, where the zero-shot translation at the inflection point is all target language tags.

\section{Mitigating Shortcut Learning}

In this section, we introduce a simple approach to alleviate the shortcut learning in MNMT (\S~\ref{sec:approach}). 
We then demonstrate that our approach improves the zero-shot performance by enhancing the generalization ability of MNMT models (\S~\ref{sec:ablation-study}). We finally validate the universality of our approach in different multilingual translation scenarios (\S~\ref{sec:main-results}).

\subsection{Approach}
\label{sec:approach}

\paragraph{Intuition.}
One straightforward way to improve zero-shot translation is to construct pseudo parallel data for all zero-shot directions~\citep{Gu:2019:ACL,Zhang2020ACL}. However, such approaches are computationally prohibitive for tasks with a large number of languages. For example, the OPUS50-En dataset consists of $49\times 48=2352$ zero-shot directions.
Another direction is to modify the model architecture~\citep{Liu2021ImprovingZT,Wu2021LanguageTM} without introducing additional training costs. Different from these directions, we propose to improve the model training to alleviate the shortcut learning. 

The starting point for our approach is an observation: NMT models suffer from catastrophic forgetting during training, where the models tend to gradually forget previously learned knowledge and swing to fit the new data that may have a different distribution~\citep{Shao:2022:ACL}. 
We can leverage the forgetting nature of model training to forget the shortcuts. 


\paragraph{Generalization Training.}

Our approach divides the training process with $N$ steps into two phases:
\begin{itemize}[leftmargin=10pt]
    \item {\em Standard Training Phase}: For the first $N-G$ training steps, we follow the standard pipeline to train the models on the full training data.
    \item {\em Generalization Training Phase}: For the last $G$ steps, we train the models only on the training example of (centric, non-centric) language pairs.
\end{itemize}
where the number of generalization training steps $G$ is a hyper-parameter. To escape the local minima of the standard training phase, we utilize a learning rate warming up at the first $0.3G$ steps and set the max learning rate to 0.0003.

\paragraph{Advantages.}
We remove the training examples of (non-centric, centric) language pairs from the generalization training phase. The potential advantages are three-fold:
\begin{enumerate}[leftmargin=10pt]
    \item It alleviates the overfitting on (non-centric, centric) language mapping, which would be forgotten by the model since they no longer occur in the later stage of model training.
    \item It enhances the role of target tags on non-centric languages, since only target sentences in non-centric languages occur in the generalization phase. Accordingly, the models can better learn to generate translation in the expected non-centric language.
    \item It potentially improves the generalization ability by enhancing the one-to-many decoding ability (e.g., one centric language to many non-centric languages), which is one criticism of MNMT models~\citep{Zhang2020ACL,Tang:2021:ACL}.
\end{enumerate}

\subsection{Ablation Study}
\label{sec:ablation-study}

In this section, we provide some insights where the generalization training improves zero-shot translation by alleviating the off-target issues.
All results are reported on the Flores validation set using the balanced CC6-En data.

\begin{figure}[h]
    \centering 
    \subfloat[w/o Pretrain]{
    \includegraphics[height=0.22\textwidth]{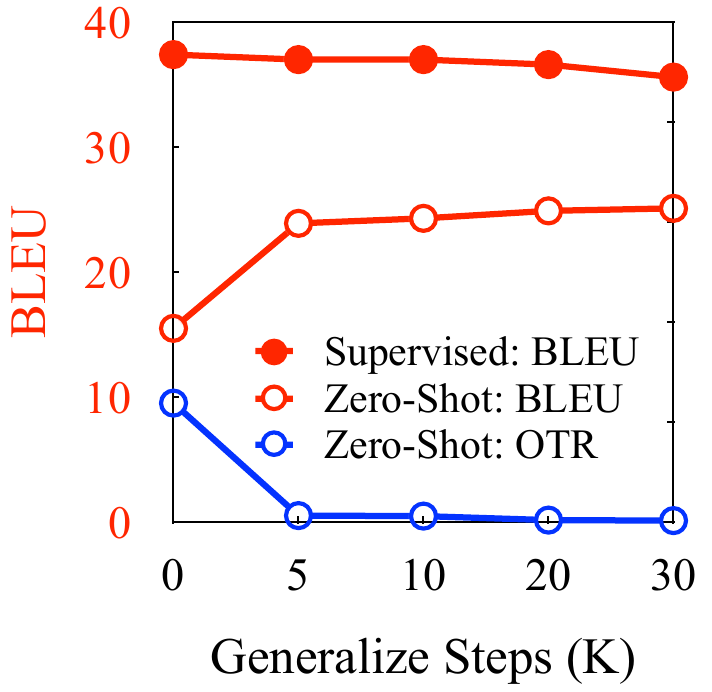}}
    \hfill
    \subfloat[w/ Pretrain]{
    \includegraphics[height=0.22\textwidth]{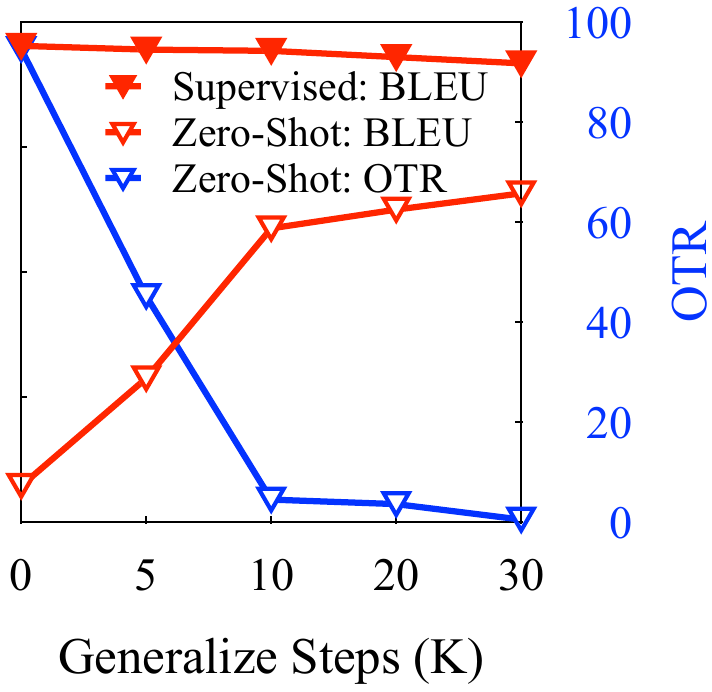}}
    \caption{Impact of generalization training steps.} 
    \label{fig:impact-of-generation-steps}
\vspace{-10pt}
\end{figure}
\paragraph{Impact of Generalization Training Steps.} 
Figure~\ref{fig:impact-of-generation-steps} shows the impact of different generalization training steps $G$. 
When $G$ increases, the performance of zero-shot translation goes up with a rapid drop of OTR. However, a large $G$ (e.g., 30K) leads to a slight performance drop for supervised translation due to the catastrophic forgetting problem. In general, our method is robust to this hyper-parameter. To balance the performances of supervised and zero-shot translation, we use G=10K in the following experiments.

\begin{figure}[h]
    \vspace{-10pt}
    \centering 
    \subfloat[w/o Pretrain]{
    \includegraphics[height=0.22\textwidth]{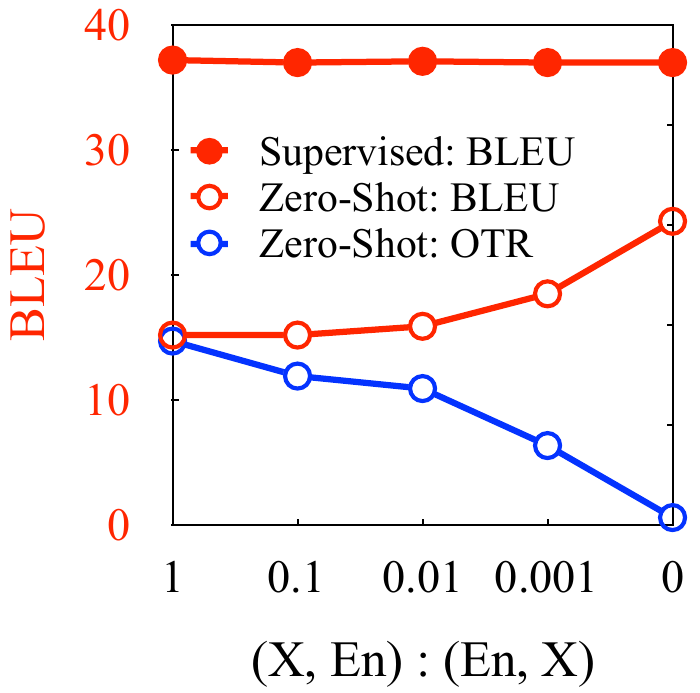}}
    \hfill
    \subfloat[w/ Pretrain]{
    \includegraphics[height=0.22\textwidth]{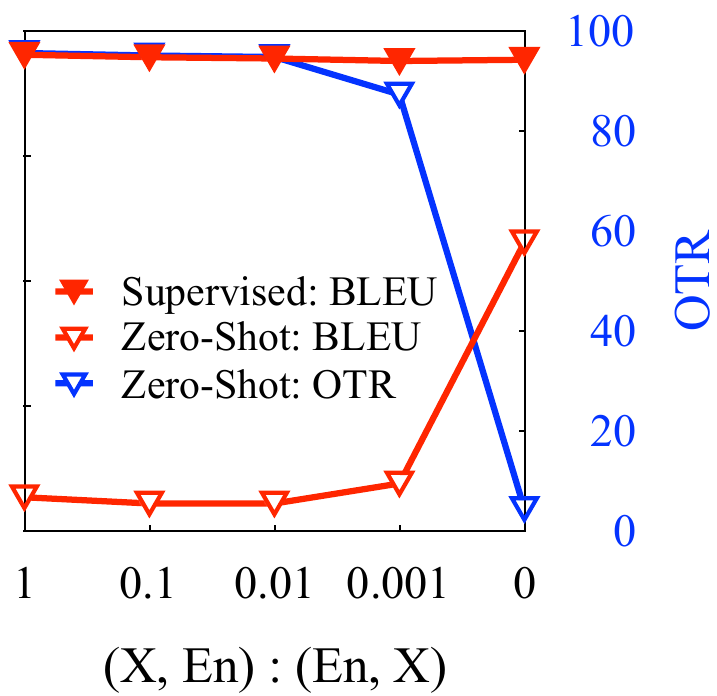}}
    \caption{Impact of different ratios of (non-centric, centric) (i.e. ``(X, En)'') and (centric, non-centric) (i.e. ``(En, X)'') instances in the generalization training stage. ``1'' denotes the vanilla training with full set of instances, and ``0'' denotes the proposed generalization training using only (centric, non-centric) instances.}
    \label{fig:impact-of-generation-instances}
\end{figure}

\paragraph{Impact of Training Instances.} Figure~\ref{fig:impact-of-generation-instances} shows the impact of (non-centric, centric) instances used in the generalization training stage. 
The off-target ratio goes down with the decreased number of used (non-centric, centric) instances, confirming the design intuition of generalization training.
However, the off-target issues are hardly mitigated for MNMT with pretraining, even when using as few as 0.1\% of (non-centric, centric) training instances. One possible reason is that MNMT with pretraining suffers from much more serious shortcut learning problem, and thus can be easier to be recalled by the corresponding instances. This finding {\em dispels the possibility of gradually reducing the number of (non-centric, centric) instances during training}.

\begin{figure}[h]
    \centering 
    \subfloat[w/o Pretrain]{
    \includegraphics[height=0.28\textwidth]{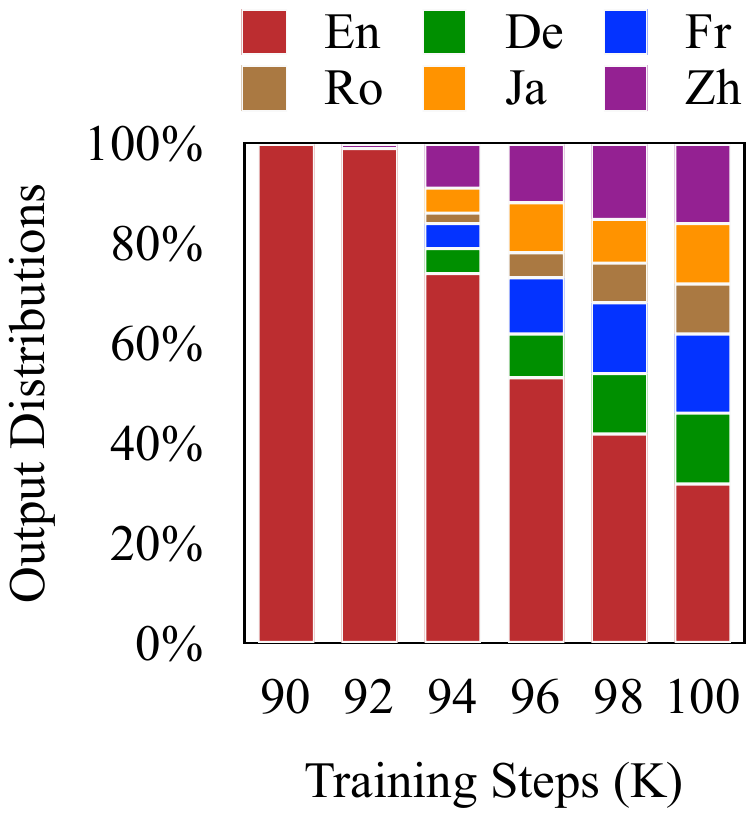}}
    \hfill
    \subfloat[w/ Pretrain]{
    \includegraphics[height=0.28\textwidth]{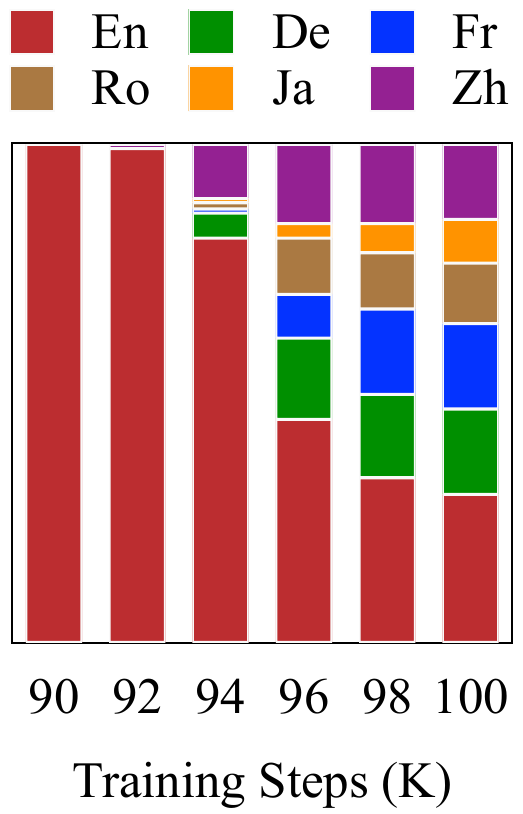}}
    \caption{Learning curves of output distributions.} 
    \label{fig:learning-curves-distributions}
\vspace{-10pt}
\end{figure}
\paragraph{Alleviating the Shortcut Learning.} 
Figure~\ref{fig:learning-curves-distributions} shows the learning curves of output distributions for translating non-centric languages {\em without target language tag}. As the generalization training phase (starting at 90K steps) progresses, MNMT models are more leaning to generate translations in non-centric languages. These results confirm our claim that our approach can alleviate the shortcut learning on mapping non-centric languages to the centric language.
Table~\ref{tab:case-ours} in Appendix shows translation examples at different generalization training steps. The off-target issues are solved with more generalization steps for MNMT model with pretraining initialization, which suffers from more severe shortcut learning problem.

\subsection{Main Results}
\label{sec:main-results}

\begin{table*}[t]
\caption{Translation performance on the Flores test set. ``All'' denotes the results on all translation directions including both supervised (e.g., 10 for CC6) and zero-shot (e.g., 20 for CC6) translation.} 
\fontsize{10}{11}\selectfont
\centering
\begin{tabular}{l  rrrr rrrr}
\toprule
\multirow{3}{*}{\bf Methods} &  \multicolumn{4}{c}{\bf  w/o Pretrain}   &    \multicolumn{4}{c}{\bf w/ Pretrain}\\
\cmidrule(lr){2-5} \cmidrule(lr){6-9}
 & \bf All  & \bf Sup.  & \multicolumn{2}{c}{\bf Zero-Shot} & \bf All  & \bf Sup.  & \multicolumn{2}{c}{\bf Zero-Shot}\\
\cmidrule(lr){2-2}\cmidrule(lr){3-3}\cmidrule(lr){4-5} \cmidrule(lr){6-6}\cmidrule(lr){7-7}\cmidrule(lr){8-9}
    & \it BLEU$\uparrow$   & \it BLEU$\uparrow$   &   \it BLEU$\uparrow$  & \it OTR$\downarrow$    & \it BLEU$\uparrow$   & \it BLEU$\uparrow$   &   \it BLEU$\uparrow$  & \it OTR$\downarrow$\\
\midrule
\multicolumn{9}{c}{\bf \em Average of Six Balanced CC6 Datasets}\\
MNMT        &  24.5 & 31.4 & 21.0 & 13.9   & 18.1 & 32.0 & 11.2 & 59.3\\
~~~+\textsc{GenTrain}  &  27.1 & 30.9 & 25.2 & 1.2    & 27.4 & 31.5 & 25.3 & 2.3\\
\midrule
\multicolumn{9}{c}{\bf \em Imbalanced CC16-En}\\
MNMT          & 17.9 & 35.0 & 15.5 & 27.6  & 7.1 & 35.9 & 3.0 & 91.4\\
~~~+\textsc{GenTrain}    & 23.8 & 34.8 & 22.2 & 1.6 & 24.0 & 35.5 & 22.4 & 2.3\\
\midrule
\multicolumn{9}{c}{\bf \em Noisy ImBalanced OPUS50-En}\\
MNMT        & 12.3  &  29.5 & 9.8 & 38.3 & 9.9 &  30.0 & 7.0 & 58.6 \\
~~~+\textsc{GenTrain}  & 17.6 &  29.2 & 15.9 & 11.4 & 18.1 &  29.9 & 16.7 &  13.4\\
\bottomrule
\end{tabular}
\label{tab:main-table}
\end{table*}

\begin{table*}[t]
\caption{Comparison with related work 
on the {balanced CC6-En dataset}.} 
\centering
\fontsize{10}{11}\selectfont
\setlength{\tabcolsep}{5pt}
\begin{tabular}{l rrrr rrrr}
\toprule
\multirow{3}{*}{\bf Methods}     &  \multicolumn{4}{c}{\bf w/o Pretrain}   &  \multicolumn{4}{c}{\bf w/ Pretrain}\\
\cmidrule(lr){2-5} \cmidrule(lr){6-9}
  &  \multicolumn{1}{c}{\bf All}  & \multicolumn{1}{c}{\bf Sup.}  & \multicolumn{2}{c}{\bf Zero-Shot}  & \multicolumn{1}{c}{\bf All} & \multicolumn{1}{c}{\bf Sup.}  & \multicolumn{2}{c}{\bf Zero-Shot}\\
\cmidrule(lr){2-2}\cmidrule(lr){3-3}\cmidrule(lr){4-5}
\cmidrule(lr){6-6}\cmidrule(lr){7-7}\cmidrule(lr){8-9}
    & \it BLEU$\uparrow$   & \it BLEU$\uparrow$   &   \it BLEU$\uparrow$  & \it OTR$\downarrow$    & \it BLEU$\uparrow$   & \it BLEU$\uparrow$   &   \it BLEU$\uparrow$  & \it OTR$\downarrow$\\
\midrule
\textsc{GenTrain}   & 28.5  &  37.0 & 24.3  & 1.5  &  28.0 &  37.7  & 23.1 &  4.4\\
\hdashline
 Residual Removing  & 25.1 & 37.2 & 19.1 & 18.5  & 14.4 & 37.8 & 2.7 & 92.6\\
 ~~~+\textsc{GenTrain}    & 29.5 & 36.8 & 25.8 & 0.2  & 29.0 & 37.5 & 24.8 & 4.1\\
\hdashline
 T-Enc Tagging            & 27.8 & 37.1 & 23.1 & 1.8 & 24.5 & 38.2 & 17.6 & 38.0\\
 ~~~+\textsc{GenTrain}    & 28.6 & 36.9 & 24.4 & 0.2 & 28.1 & 37.8 & 23.2 & 5.2\\
\bottomrule
\end{tabular}
\label{tab:comparison}
\end{table*}

In this section, we validate the effectiveness of our approach on different MNMT benchmarks.

\paragraph{MNMT Benchmarks.}
Table~\ref{tab:main-table} lists the translation results on different training datasets to simulate different MNMT scenarios, including different language distributions (balanced and imbalanced), different number of languages (6, 16, 50), and dataset with noise.  
Clearly, our approach consistently and significantly improves zero-shot translation in all cases, demonstrating the robustness of the proposed generalization training approach.

\begin{table}[h]
\caption{Results of MNMT models trained on the CC6-En balanced dataset measured by other evaluation metrics.}
\setlength{\tabcolsep}{3pt}
\centering
\begin{tabular}{l rr rr}
\toprule
\multirow{2}{*}{\bf Model} 
& \multicolumn{2}{c}{\bf Supervised}  & \multicolumn{2}{c}{\bf Zero-Shot}\\
\cmidrule(lr){2-3}\cmidrule(lr){4-5}
    &   \bf COMET$\uparrow$  & \bf ChrF$\uparrow$    &   \bf COMET$\uparrow$ & \bf ChrF$\uparrow$   \\
\midrule
\multicolumn{5}{c}{\bf  w/o Pretrain}\\
Vanilla &  0.646 & 57.11 & -0.053& 33.93\\
~~~+\textsc{GenTrain}  &0.642  & 56.93  & 0.389 & 40.88\\
\midrule
\multicolumn{5}{c}{\bf w/ Pretrain}\\
Vanilla         & 0.665 & 57.58  & -0.544 & 16.30\\
~~~+\textsc{GenTrain}   & 0.661 & 57.21  & 0.322 & 38.05\\
\bottomrule
\end{tabular}
\label{tab:main-comet-chrf}
\end{table}

Table~\ref{tab:main-comet-chrf} lists the detailed results on the CC6-En balanced datasets with COMET and ChrF evaluation metrics. These results demonstrate that our approach can significantly improve the zero-shot performance.

\begin{table}[t]
\caption{Supervised translation performance on the test set for the six {\bf balanced CC6 datasets}. ``X$\rightarrow$C'' denotes the results of non-centric to centric supervised translation, and  ``C$\rightarrow$X'' denotes the results in reverse direction.}
\centering
\begin{tabular}{l  rr rr}
\toprule
\multirow{2}{*}{\bf Methods} &  \multicolumn{2}{c}{\bf  w/o Pretrain}   &    \multicolumn{2}{c}{\bf w/ Pretrain}\\
\cmidrule(lr){2-3} \cmidrule(lr){4-5}
  & \bf X$\rightarrow$C  & \bf C$\rightarrow$X  & \bf X$\rightarrow$C  & \bf C$\rightarrow$X\\
\midrule
MNMT        &   31.8  &   31.0 &  32.4 & 31.5\\
~~~+\textsc{GenTrain}  &  31.1 & 31.1 & 31.7 & 31.6\\
\bottomrule
\end{tabular}
\label{tab:supervised-translation}
\end{table}

Table~\ref{tab:supervised-translation} lists the results of supervised translation in two separate directions: non-centric to centric, and centric to non-centric.  The marginal performance
decline of supervised translation is mainly from the translation from non-centric languages to centric languages. This is intuitive, since the training examples on these directions are not presented in the later training stage, thus some useful information along with the overfitted language mapping patterns are forgotten by MNMT models. 

\paragraph{Comparison with Relevant Works.}
We also compare two strong baselines: 
\begin{itemize}
    \setlength\itemsep{0pt}
    \item {\em Residual Removing}~\citep{Liu2021ImprovingZT}: removing the residual connection on an encoder layer.
    \item {\em T-Enc Tagging}~\citep{Wu2021LanguageTM}: only attaching the target tag to the beginning of encoder input.
\end{itemize}
Both methods have empirically shown improvement on zero-shot translation, while {we enhance the understanding of them from a shortcut learning perspective.}
Our study provides an explanation: (1) Residual Removing mitigates the shortcut language mapping by reducing the dependency on low-level features of source language tags; (2) T-Enc Tagging retains the flexibility by only specifying the target languages, thus breaks away from the shortcuts (source tag, target tag).
Table~\ref{tab:comparison} lists the results. Our approach outperforms both strong baselines when using individually, and the improvement is more significant for MNMT with pretraining.
Combining them together can further improve the zero-shot performance. This is intuitive, since the two strong baselines improve the model architecture and our approach reforms the model training. 

\begin{table*}[t!]
\caption{{\bf Translation Examples} from the models trained on balanced CC6-En dataset at different training steps.}
\centering
\resizebox{0.9\textwidth}{!}{
\begin{tabular}{cc p{8cm}p{8cm} }
\toprule
\bf Steps & \bf Pre-  & \multicolumn{1}{c}{\bf Supervised (Fr-En)}  & \multicolumn{1}{c}{\bf Zero-Shot (Fr-Zh)} \\
\cmidrule(lr){1-1} \cmidrule(lr){2-2} \cmidrule(lr){3-3} \cmidrule(lr){4-4}
 \bf Src.  & \bf Train & C'est une bonne occasion d'admirer les aurores boréales, car le ciel sera sombre pratiquement toute la journée. & C'est une bonne occasion d'admirer les aurores boréales, car le ciel sera sombre pratiquement toute la journée.\\
\bf Ref.  & &This offers a good opportunity to see the Aurora borealis, as the sky will be dark more or less around the clock. & \begin{CJK}{UTF8}{gkai} 这提供了一个可以看到北极光的绝佳机会,因为天空将或多或少连续一整天都是暗的。\end{CJK} \\
\hline
\multirow{2}{*}{\bf 100}  & \texttimes  & , , , , , . &   \_\_en\_XX\_\_ , , , , ,\\
                         & \checkmark  & C'est une bonne occasion d'admirer les aurores boréales, car le ciel sera sombre pratiquement toute la journée & C'est une bonne occasion d'admirer les aurores boréales, car le ciel sera sombre pratiquement toute la journée\\
\hdashline
\multirow{2}{*}{\bf 200}  & \texttimes  &  The the the the the the the the. &  of the the the the the the. \\
                         & \checkmark  &  C'est a good opportunity to see the aurores boréales, because the moon will be dark for most of the day. & C'est une bonne occasion de voir les aurores boréales, car le ciel sera sombre pratiquement toute la journée. \\
\hdashline
\multirow{2}{*}{\bf 300}  & \texttimes  & The is a a a a a a to the. & of the , and the , and the .\\
                         & \checkmark  & This is a good opportunity to see the borrowing auroras, because the sky will be dark most of the day. &\begin{CJK}{UTF8}{gkai} 这是一个好机会,欣赏夕阳,因为天空会变得暗彻夜。 \end{CJK} \\
\hdashline
\multirow{2}{*}{\bf 400}  & \texttimes  & It’s are not not not not not. & “ “ “ “ “ “ “ “, and I’t you you you you can be be be the world. \\
                         & \checkmark  & It’s a good time to see boring auroras, because the sky will be dark for most of the day.& \begin{CJK}{UTF8}{gkai} 这是值得观赏的夕阳,因为白天的天空将很暗。 \end{CJK}  \\
\hdashline
\multirow{2}{*}{\bf 500}  & \texttimes  & It’s the world’s the world’s the world. & \begin{CJK}{UTF8}{gkai} 是是是是是是是是是是是。 \end{CJK}   \\
                         & \checkmark  & This is a good opportunity to see boring auroras, because the sky will be dark almost all day. & \begin{CJK}{UTF8}{gkai} 这是一个很好的机会,看看闪烁的夕阳,因为天空将很暗整个
晚上。 \end{CJK}  \\
\hdashline
\multirow{2}{*}{\bf 600}  & \texttimes  & It is the world of the world, the world’s the world’s the world. & \begin{CJK}{UTF8}{gkai} 在在在在在在在在在在在在在在在在的。 \end{CJK} \\
                         & \checkmark  &   & \_\_en\_XX\_\_  \_\_en\_XX\_\_ \_\_en\_XX\_\_ \_\_en\_XX\_\_ \_\_en\_XX\_\_\\
\hdashline
\multirow{2}{*}{\bf 700}  & \texttimes  & It is a few years of the same time, but it is the same time. & \begin{CJK}{UTF8}{gkai} 他们他们他们他们他们他们的。\end{CJK}  \\
                         & \checkmark  & This is a good opportunity to see boring auroras, because the sky will be dark almost all day. & This is a good opportunity to see boring auroras, because the sky will be dark almost all day. \\
\hdashline
\multirow{2}{*}{\bf 800}  & \texttimes  & It is a lot of the same time, the same time of the same time of the world.& \begin{CJK}{UTF8}{gkai} 它,我们我们我们我们我们我们我们的。 \end{CJK}  \\
                            & \checkmark  & This is a good opportunity to admire the boring auroras because the sky will be dark almost all day. &This is a good opportunity to admire the boréal auroras, because the sky will be dark practically all day. \\
\hdashline
\multirow{2}{*}{\bf 900}  & \texttimes  & It’s a lot of the same time, it is a lot of the same time. & \begin{CJK}{UTF8}{gkai} 因为因为他们他们他们他们\end{CJK}\\
                         & \checkmark  & This is a good opportunity to admire the boréal auroras, because the sky will be dark practically all day. & This is a good opportunity to admire the boréal auroras, because the sky will be dark practically all day. \\
\hdashline
\multirow{2}{*}{\bf 1k}  & \texttimes  & It’s a lot of the same time, it’s a lot of the same time, it’s very important. &  \begin{CJK}{UTF8}{gkai} 因此,它它它它,它它它它它。 \end{CJK}  \\
                            & \checkmark  & It is a good opportunity to admire the bored auroras, because the sky will be dark almost all day. & It is a good opportunity to admire the bored auroras, because the sky will be dark almost all day. \\
\hdashline
\multirow{2}{*}{\bf 10k}  & \texttimes  &  It is a good opportunity to admire the Goldenores, because the sky will be dark almost all day. & \begin{CJK}{UTF8}{gkai} 这是一个很好的机会, \end{CJK}  admire the boreal aurores, because the sky will be dark almost all day. \\
                            & \checkmark   & This is a good opportunity to admire the boreal auroras, because the sky will be dark almost all day.  & This is a good opportunity to admire the boreal aurores, because the sky will be dark virtually all day. \\
\hdashline
\multirow{2}{*}{\bf 50k}  & \texttimes  &  This is a good opportunity to admire the aurorae, because the sky will be dark almost all day. & \begin{CJK}{UTF8}{gkai} 这是一个很好的机会来欣赏北极极光,因为天几乎会整天黑暗。 \end{CJK} \\
                            & \checkmark   &  This is a good opportunity to admire the Northern Lights, as the sky will be dark practically all day long. &  This is a good opportunity to admire the Northern Lights, because the sky will be dark practically all day long. \\
\hdashline
\multirow{2}{*}{\bf 100k}  & \texttimes  & This is a good opportunity to admire the Northern Lights, as the sky will be dark almost all day. & It is a good opportunity to admire the boreal aurores, because the sky will be dark almost all day. \\
                            & \checkmark   & This is a good opportunity to admire the Northern Lights, as the sky will be dark almost all day. & This is a good opportunity to admire the Northern Lights, as the sky will be dark practically all day. \\
\bottomrule
\end{tabular}
}
\label{tab:case-vanilla}
\end{table*}

Tabel~\ref{tab:case-vanilla} shows the translation examples by the models at different training steps. We randomly select a French sentence and translate it into English (supervised direction) and Chinese (zero-shot direction) by both the MNMT models finetuned from mBART50 and training from scratch. These translation examples are consistent with our findings such that:
\begin{itemize}
    \item Off-target translations are in the centric language. For example, the generated translation of Fr-Zh at step 100k is in English, which is the centric language of the CC6-En dataset. 
    \item Off-target translations occur at the late training stage. Both the model trained from scratch and finetuned from mBART50 can generate in-target translation sentences at the early stage of training. For example, the generated translation of Fr-Zh at 500 steps is in Chinese. 
    \item Finetuning from mBART50 will accelerate the learning of not only supervised translation but also the shortcuts. Compared with the model trained from scratch, the model finetuned from mBART50 can generate more fluent sentences at the early stage of training (e.g., example at step 500). However, it learns the shortcut of (non-central, central) mapping at step 700, which is much earlier than the model trained from scratch (i.e., after 10k steps).
    \item The MNMT model finetuned from mBART50 shows a transition process of shortcuts from the copy behavior to (non-central, central) mapping. At step 100, the model finetuned from mBART50 only copies the source sentence. But, after the inflection point at around step 600, the model starts to generate sentences into English, which is the centric language.
\end{itemize}

\begin{table*}[t!]
\caption{{\bf Translation Examples} from the models trained on balanced CC6-En dataset {\bf with the proposed approach} at different training steps.}
\centering
\resizebox{0.95\textwidth}{!}{
\begin{tabular}{cc p{8cm}p{8cm}}
\toprule
\bf Steps & \bf Pre-  & \multicolumn{1}{c}{\bf Zero-Shot (Fr-Zh)} &  \multicolumn{1}{c}{\bf Zero-Shot (Fr-Zh)}\\
\cmidrule(lr){1-1} \cmidrule(lr){2-2}  \cmidrule(lr){3-3}  \cmidrule(lr){4-4} 
 \bf Src.  & \bf Train &  C'est une bonne occasion d'admirer les aurores boréales, car le ciel sera sombre pratiquement toute la journée. & Des éléments comme le calcium et le potassium sont considérés comme des métaux. Bien sûr, il y a aussi des métaux comme l’argent et l’or.\\
\bf Ref.  & & \begin{CJK}{UTF8}{gkai} 这提供了一个可以看到北极光的绝佳机会,因为天空将或多或少连续一整天都是暗的。\end{CJK} &   \begin{CJK}{UTF8}{gkai} 钙、钾等元素属于金属,银和金等元素当然也是金属。\end{CJK} \\
\hline
\multirow{2}{*}{\bf 0k}  & \texttimes  &  It is a good opportunity to admire the boreal aurores, because the sky will be dark almost all day.  & \begin{CJK}{UTF8}{gkai} 元素 \end{CJK} such as calcium and potassium are considered metals. of course, there are also other metals such as silver and gold. \\
                         & \checkmark  & This is a good opportunity to admire the Northern Lights, as the sky will be dark practically all day. &  Elements like calcium and potassium are considered as metals. Of course, there are also metals like silver and gold.\\
\hdashline
\multirow{2}{*}{\bf 2k}  & \texttimes  &  \begin{CJK}{UTF8}{gkai} 这是欣赏北极光的绝佳机会,因为天将几乎全天阴暗。\end{CJK} & \begin{CJK}{UTF8}{gkai}元素\end{CJK}s such as calcium and potassium are considered metals. of course, there are other metals such as silver and gold. \\
                         & \checkmark  &   \begin{CJK}{UTF8}{gkai} 这是观赏北极光的绝佳机会,因为天几乎整天都会阴暗。 \end{CJK} &Elements like calcium and potassium are considered as metals. Of course, there are also metals like silver and gold. \\
\hdashline
\multirow{2}{*}{\bf 4k}  & \texttimes  & \begin{CJK}{UTF8}{gkai} 这是欣赏北极光的好机会,因为天空将几乎全天阴暗。 \end{CJK} & \begin{CJK}{UTF8}{gkai} 像钙和钾这样的元素被视为金属,当然还有其他金属,如银和金。\end{CJK}  \\
                         & \checkmark  &  \begin{CJK}{UTF8}{gkai} 这是观赏北极光的一个绝佳机会,因为天几乎整天都会阴暗。 \end{CJK} &  Calcium and potassium are considered metals. \begin{CJK}{UTF8}{gkai} 当然,\end{CJK}  there are also metals like silver and gold. \\
\hdashline
\multirow{2}{*}{\bf 6k}  & \texttimes  & \begin{CJK}{UTF8}{gkai} 这是欣赏北极光的好机会,因为天空将几乎全天阴暗。 \end{CJK} & \begin{CJK}{UTF8}{gkai} 像钙和钾这样的元素被视为金属,当然还有其他金属,如银和黄金。\end{CJK} \\
                         & \checkmark  & \begin{CJK}{UTF8}{gkai} 这是观赏北极光的绝佳机会,因为天空几乎全天都是阴暗的。 \end{CJK} & Calcium and potassium are considered metals.\begin{CJK}{UTF8}{gkai} 当然\end{CJK}, silver and gold are also considered metals. \\
\hdashline
\multirow{2}{*}{\bf 8k}  & \texttimes  & \begin{CJK}{UTF8}{gkai} 这是欣赏北极光的好机会,因为天将是黑几乎一整天。 \end{CJK} & \begin{CJK}{UTF8}{gkai} 像钙和钾这样的元素被视为金属,当然还有其他金属,如银和黄金。\end{CJK} \\
                            & \checkmark  & \begin{CJK}{UTF8}{gkai} 这是观赏北极光的好时机,因为天几乎整天都会阴暗。 \end{CJK} & Calcium and potassium are considered metals. \begin{CJK}{UTF8}{gkai} 当然\end{CJK}, there are also metals like silver and gold.\\
\hdashline
\multirow{2}{*}{\bf 10k}  & \texttimes  & \begin{CJK}{UTF8}{gkai} 这是欣赏北极光的好机会,因为天将是黑几乎一整天。\end{CJK} &  \begin{CJK}{UTF8}{gkai} 像钙和钾这样的元素被视为金属,当然还有其他金属,如银和黄金。\end{CJK} \\
                            & \checkmark  & \begin{CJK}{UTF8}{gkai} 这是观赏北极光的好机会,因为天空几乎全天都是阴暗的。 \end{CJK}  & \begin{CJK}{UTF8}{gkai} 钙和钾等元素被认为是金属,当然还有其他金属,如银和黄金。\end{CJK}\\
\bottomrule
\end{tabular}
}
\label{tab:case-ours}
\end{table*}

Tabel~\ref{tab:case-ours} shows the translation examples by the models trained with our method at different generalization training steps. We randomly select two French sentences and translate them into Chinese (zero-shot direction) by both the MNMT models finetuned from mBART50 and trained from scratch. These cases show that our generalization training method can help MNMT models quickly forget the learned shortcuts of erroneous language mapping (to English) and generate in-target translations (to Chinese).

\section{Conclusion}

We connect the commonly-cited off-target issues in MNMT to a shortcut learning on the supervised language mapping. 
We also identify and explain a critical side-effect of multilingual pretraining for MNMT: 
pretraining initialization introduces remarkably more off-target issues by accelerating and aggravating the shortcut learning.
Based on these findings, we propose a simple and effective training strategy to improve the zero-shot translation performance by mitigating the shortcut learning without introducing any computational cost.

\section*{Acknowledgment}
The work described in this paper was supported by the Research Grants Council of the Hong Kong Special Administrative Region, China (No. CUHK 14206921 of the General Research Fund).


\section{Appendix}

\subsection{Model Training Results on Balanced CC6-Ro Data}

 Figure~\ref{fig:learning-curves-vanilla-ro} shows the learning curves of models trained on balanced CC6-Ro data. Similar to the results on balanced CC6-En data, MNMT models keep improving the performance of supervised translation, but sacrifice the generalization ability on zero-shot translation.

\begin{figure*}[h]
    \vspace{-10pt}
    \centering 
    \subfloat[BLEU: w/o Pretrain]{
    \includegraphics[height=0.24\textwidth]{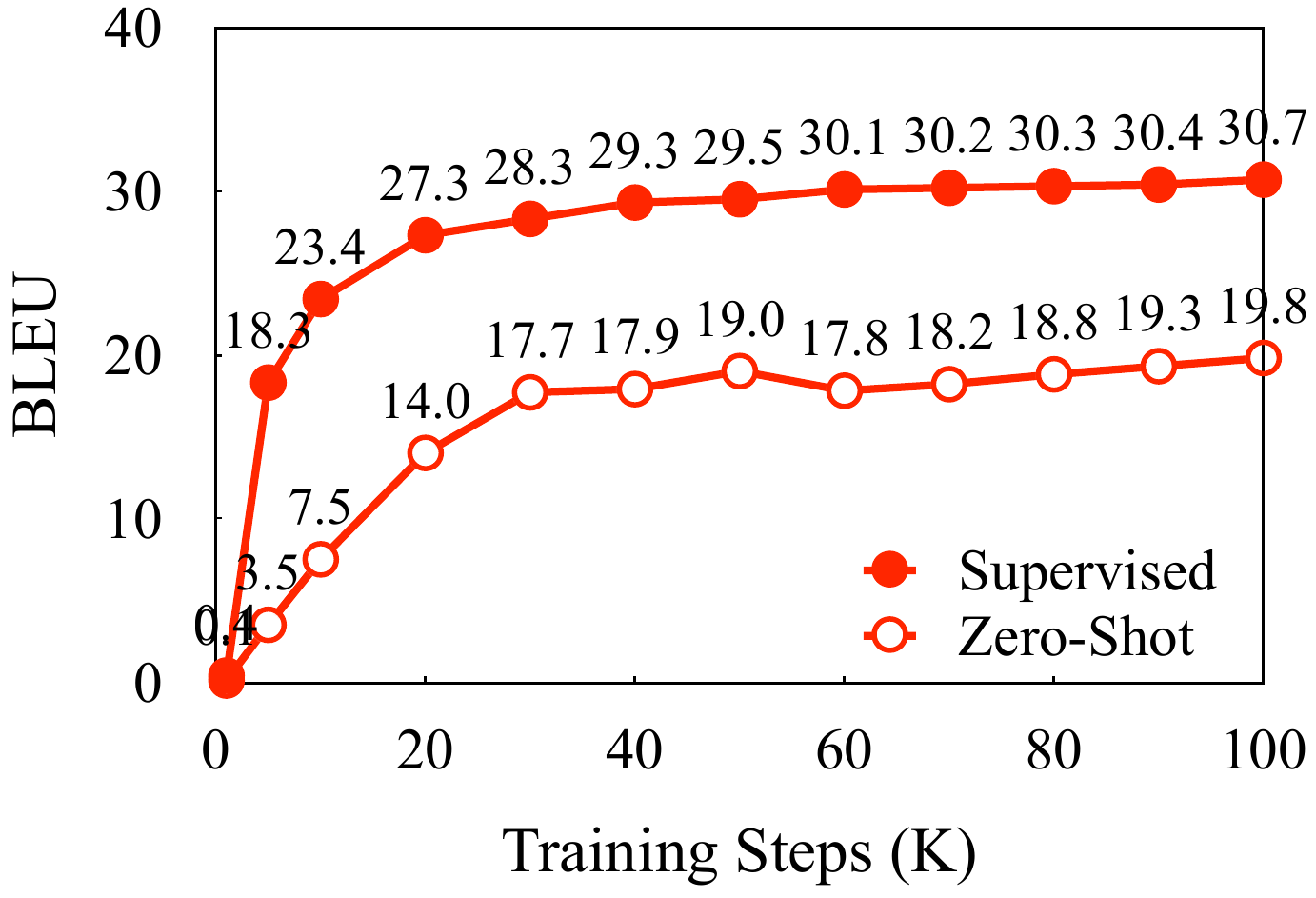}}
    \hspace{0.07\textwidth}
    \subfloat[OTR: w/o Pretrain]{
    \includegraphics[height=0.24\textwidth]{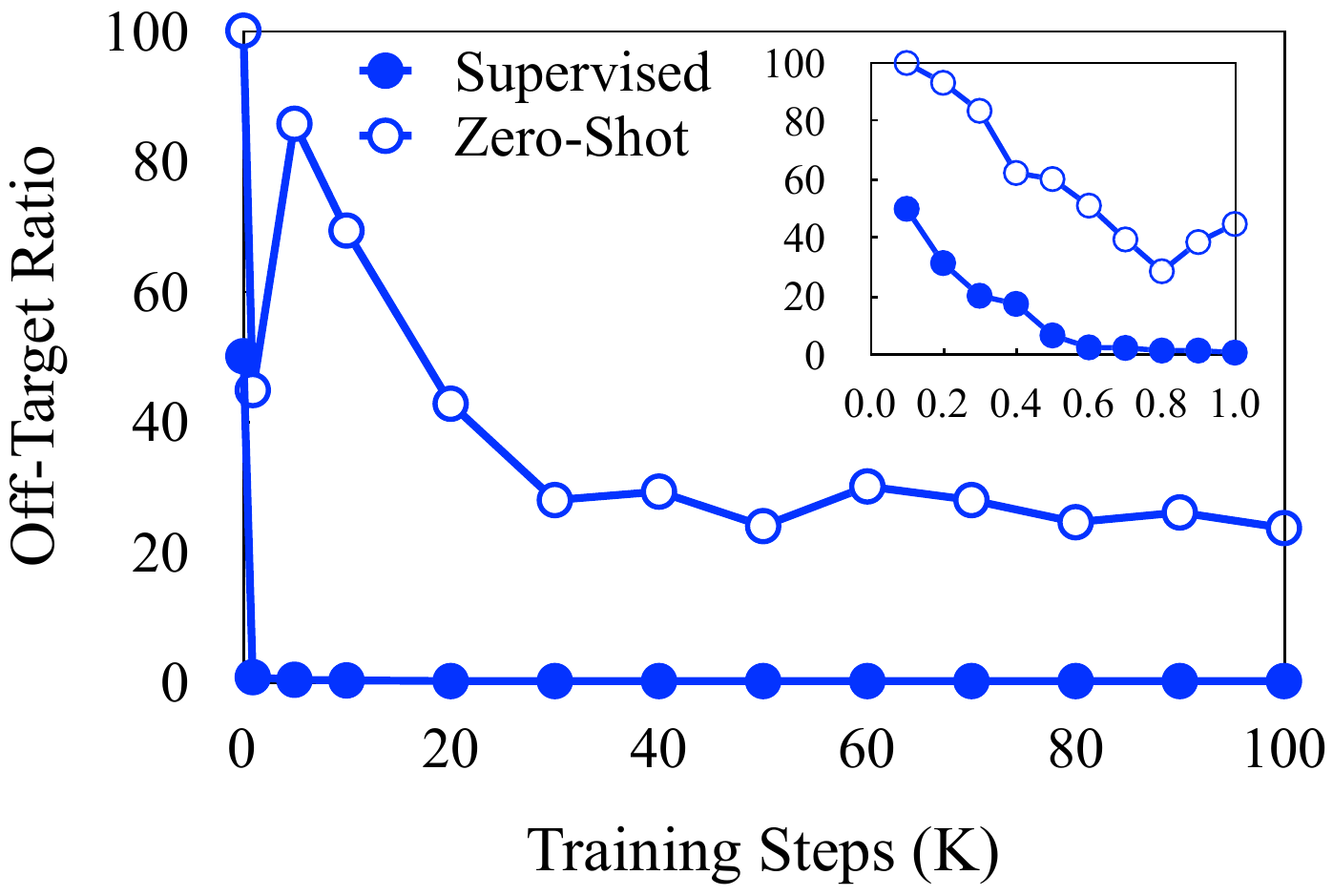}}
    \\
    \subfloat[BLEU: w/ Pretrain]{
    \includegraphics[height=0.24\textwidth]{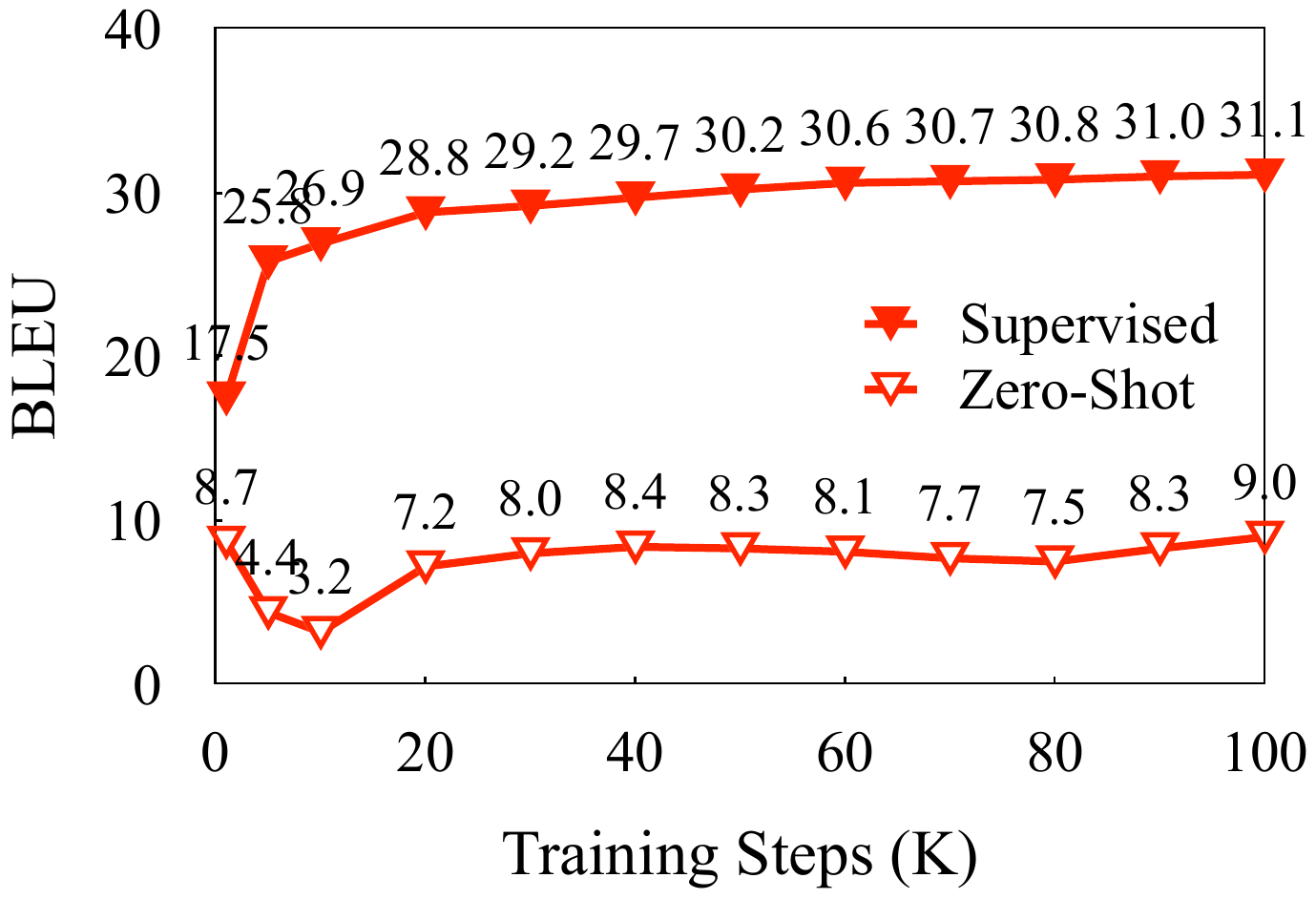}}
    \hspace{0.07\textwidth}
    \subfloat[OTR: w/ Pretrain]{
    \includegraphics[height=0.24\textwidth]{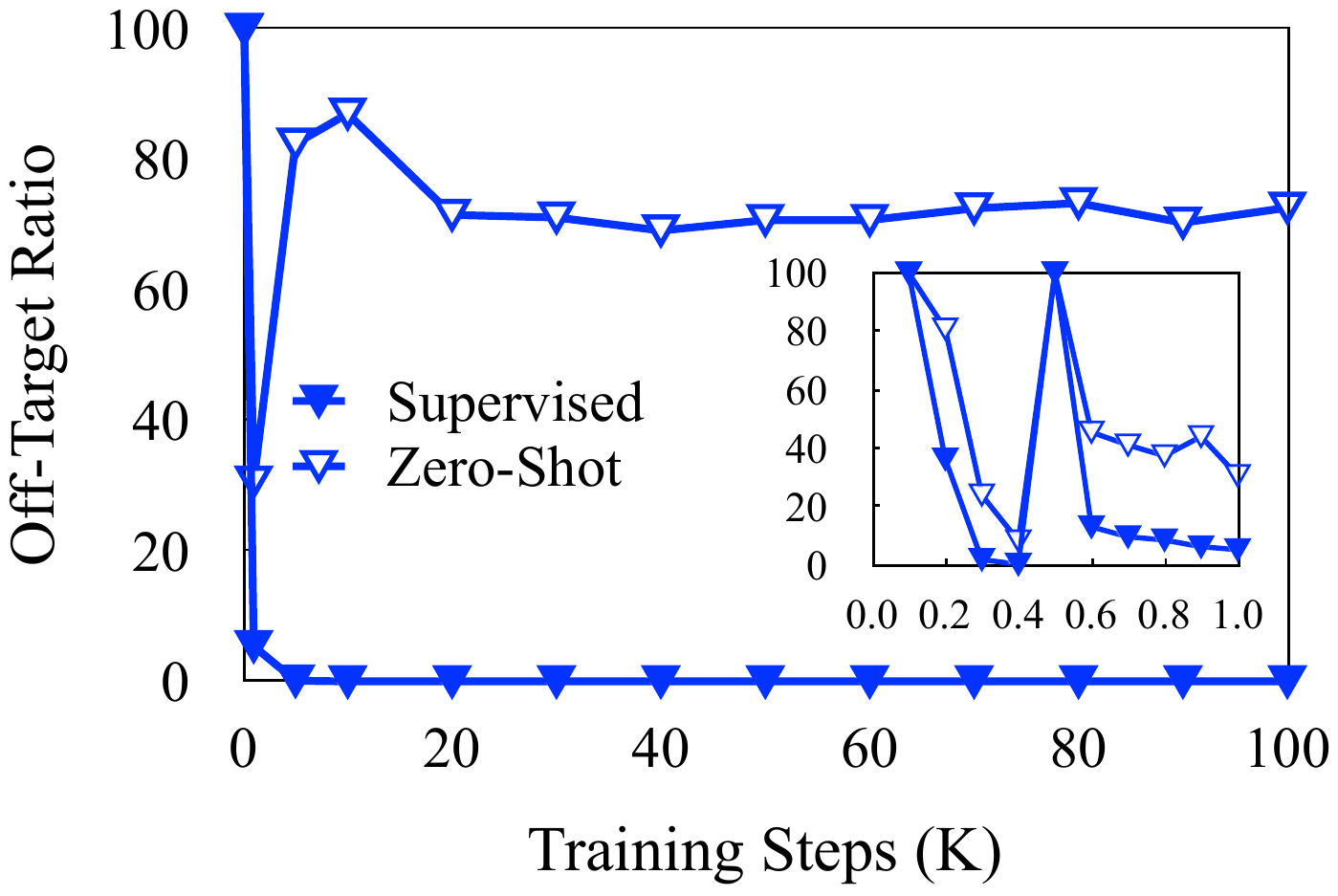}}
    \caption{Learning curves of the MNMT model ({\bf balanced CC6-Ro data}) on the validation set.}
    \label{fig:learning-curves-vanilla-ro}
\end{figure*}

\subsection{Experimental Results on Small Scale Model Architectures}
\label{app:smaller-arch}

\begin{table}[h]
\caption{Translation performance of {\bf Transformer-Base} model on the Flores test set. ``All'' denotes the results on all translation directions including both supervised (e.g., 10 for CC6) and zero-shot (e.g., 20 for CC6) translation. }
\fontsize{9}{10}\selectfont
\setlength{\tabcolsep}{5pt}
\centering
\begin{tabular}{l rrrr}
\toprule
\multirow{2}{*}{\bf Methods} & \bf All  & \bf Sup.  & \multicolumn{2}{c}{\bf Zero-Shot}\\
\cmidrule(lr){2-2}\cmidrule(lr){3-3}\cmidrule(lr){4-5}
    & \it BLEU$\uparrow$   & \it BLEU$\uparrow$   &   \it BLEU$\uparrow$  & \it OTR$\downarrow$\\
\midrule
\multicolumn{5}{c}{\bf Balanced CC6-En}\\
MNMT        &  19.9 & 31.6 &14.1  &  36.1\\
~~~+\textsc{GenTrain}  & 23.9 & 31.4 & 20.1 &  1.8\\
\midrule
\multicolumn{5}{c}{\bf Imbalanced CC6-En}\\
MNMT        &  19.8 & 30.9 &14.3 & 35.4 \\
~~~+\textsc{GenTrain}  & 22.6 & 30.7 & 18.6 &  2.4\\
\midrule
\multicolumn{5}{c}{\bf OPUS-100}\\
MNMT        & 11.7  & 27.0 & 4.8 & 45.0 \\
~~~+\textsc{GenTrain}  & 16.3 & 26.8 & 11.5 &  13.3\\
\bottomrule
\end{tabular}
\label{tab:small}
\end{table}

To validate our generalization training method in different model architecture, we conduct the experiments on Transformer-base architecture, which consists of 6 encoder layers and 6 decoder layers with 512 dimensions. Table~\ref{tab:small} shows the translation performance of the model trained on balanced and imbalance CC6 English dataset without and with our method. The conclusions still hold such that a) off-target translations are mainly in the centric language, b)Our method can improve the zero-shot translation performance and alleviate the off-target issues.

\subsection{Detailed Results on Balanced CC6 Datasets}

\begin{table*}[t]
\caption{Translation performance on the test set for the six {\bf balanced CC6 datasets}. ``All'' denotes the results on all translation directions including both supervised (e.g., 10 directions) and zero-shot (e.g., 20 directions) translation.}
\setlength{\tabcolsep}{5pt}
\centering
\begin{tabular}{l  rrrr rrrr}
\toprule
\multirow{3}{*}{\bf Methods} &  \multicolumn{4}{c}{\bf  w/o Pretrain}   &    \multicolumn{4}{c}{\bf w/ Pretrain}\\
\cmidrule(lr){2-5} \cmidrule(lr){6-9}
\multirow{2}{*}{\bf Methods} & \bf All  & \bf Sup.  & \multicolumn{2}{c}{\bf Zero-Shot} & \bf All  & \bf Sup.  & \multicolumn{2}{c}{\bf Zero-Shot}\\
\cmidrule(lr){2-2}\cmidrule(lr){3-3}\cmidrule(lr){4-5}
\cmidrule(lr){6-6}\cmidrule(lr){7-7}\cmidrule(lr){8-9}
    & \it BLEU$\uparrow$   & \it BLEU$\uparrow$   &   \it BLEU$\uparrow$  & \it OTR$\downarrow$    & \it BLEU$\uparrow$   & \it BLEU$\uparrow$   &   \it BLEU$\uparrow$  & \it OTR$\downarrow$\\
\midrule
\multicolumn{9}{c}{\bf \em Balanced CC6-En}\\
MNMT        &  22.5 &   37.2    &   15.2    &   36.8    & 14.5  &   38.1    &   2.7     &   95.6\\
~~~+\textsc{GenTrain}  & 28.5  &   37.0    &   24.3    &   1.5     & 28.0  &   37.7    &   23.1    &   4.4\\
\midrule
\multicolumn{9}{c}{\bf \em Balanced CC6-De}\\
MNMT        &  26.1 &   29.6    &   24.4    &   8.5     & 11.7  &   30.2    &   2.5     &   95.8\\
~~~+\textsc{GenTrain}  &  28.1 &   29.3    &   27.5    &   0.7     &  28.1 &   29.8    &   27.2    &   2.4\\
\midrule
\multicolumn{9}{c}{\bf \em Balanced CC6-Zh}\\
MNMT        & 26.6  &   29.4    &   25.2    &   4.1     & 24.2  &   30.0    &   21.3    &   21.0\\
~~~+\textsc{GenTrain}  & 27.7  &   29.0    &   27.1    &   0.4     &  28.5 &   29.5    &   28.0    &   0.6\\
\midrule
\multicolumn{9}{c}{\bf \em Balanced CC6-Ro}\\
MNMT        &  22.9 &   30.5    &   19.1    &   24.3    & 15.9  &   30.8    &   8.4     &   73.8\\
~~~+\textsc{GenTrain}  & 26.3  &   30.0    &   24.5    &   0.9     &  26.9 &   30.3    &   25.2    &   0.9\\
\midrule
\multicolumn{9}{c}{\bf \em Balanced CC6-Fr}\\
MNMT        &  26.8 &   33.1    &   23.6    &   7.0     &  23.1 &   33.6    &   17.9    &   34.4\\
~~~+\textsc{GenTrain}  & 28.2  &   32.6    &   26.0    &   0.8     & 29.2  &   33.3    &   27.1    &   0.7\\
\midrule
\multicolumn{9}{c}{\bf \em Balanced CC6-Ja}\\
MNMT        & 21.9  &   28.5    &   18.6    &   23.4    &  19.3 &   29.3    &   14.3    &   49.2\\
~~~+\textsc{GenTrain}  &  23.2 &   28.1    &   20.8    &   5.1     & 23.6  &   28.9    &   20.9    &   5.3\\
\bottomrule
\end{tabular}
\label{tab:main-table-balanced}
\end{table*}

Table~\ref{tab:main-table-balanced} lists the detailed results on the balanced datasets with different centric languages.

\subsection{Large Language Models and Off-Target Issues}

Large Language Models, such as ChatGPT and Gemini, have been proposed and developed recently. Previous works have evaluated their translation performance in both bi-lingual setting~\citep{Jiao2023IsCA} and multilingual setting~\citep{Zhu2023MultilingualMT}. According to the report\citep{Wu2024AdaptingLL}, LLMs, such as GPT-4, still suffer from off-target issues. This can be attributed to the dominance of the English corpus in the training and alignment data~\citep{Achiam2023GPT4TR}, which also belongs to shortcut learning from a data perspective.

\subsection{Multiple Centric Languages Setting Does not Suffer From Off-Target Issues}

We conduct experiments on datasets with multiple centric languages (e.g., ``En+De'' in Table~\ref{tab:centric-langs-one}), where the language mapping patterns of (non-centric, centric) are more complex and thus are difficult to overfit. For example, sentences in French are translated into two different centric languages (e.g., English and German for the ``En+De'' data). As listed in Table~\ref{tab:centric-langs-one}, off-target issues never occur on datasets with multiple centric languages. However, this setting is not practical due to the lack of training data. Based on this observation, we propose a novel and data effecient method to mitigate the off-target issue without the need of any multiple centric languages data.

\begin{table}[t]
\caption{Results on {\bf balanced CC6 datasets} with multiple centric languages.}
\begin{tabular}{cc rr rr}
\toprule
\bf Cen. & {\bf Pre-} & \multicolumn{1}{c}{\bf Sup.}  & \multicolumn{3}{c}{\bf Zero-Shot}\\
\cmidrule(lr){3-3}\cmidrule(lr){4-6}
\bf Lang.   & \bf Train & \it BLEU   &   \it BLEU  & \it OTR & \it OTR$_C$\\
\cmidrule(lr){1-2}\cmidrule(lr){3-6}
\multirow{2}{*}{\bf En+De}  & \texttimes  & 33.4 & 29.3 & 0.1 & 0.1 \\
                         & \checkmark  & 34.0 & 29.7 & 0.1 & 0.0 \\
\hdashline
\multirow{2}{*}{\bf En+Zh}  & \texttimes  & 32.7 & 29.0 & 0.2 & 0.1  \\
    & \checkmark   &  33.3 & 29.1 & 0.4 & 0.3 \\
\hdashline
\multirow{2}{*}{\bf De+Zh}  & \texttimes  & 30.3 & 33.2 & 0.2 & 0.1 \\
    & \checkmark   &   30.5 & 33.5 & 0.1 & 0.1  \\
\bottomrule
\end{tabular}
\label{tab:centric-langs-two}
\end{table}



\bibliographystyle{cas-model2-names}

\bibliography{cas-refs}


\clearpage

\bio{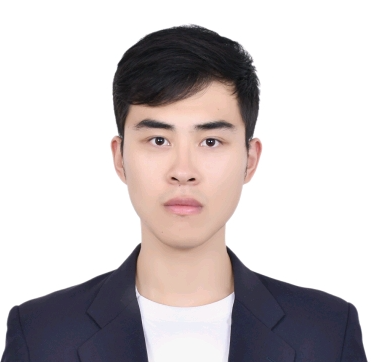}
Wenxuan Wang is a Ph.D. candidate at the Department of Computer Science and Engineering in The Chinese University of Hong Kong, advised by Prof. Michael R. Lyu. He received his B.S. from Huazhong University of Science and Technology in 2017. His research interests are mainly in the reliability of natural language processing models and software, such as machine translation models and pre-training language models.
\endbio

\bio{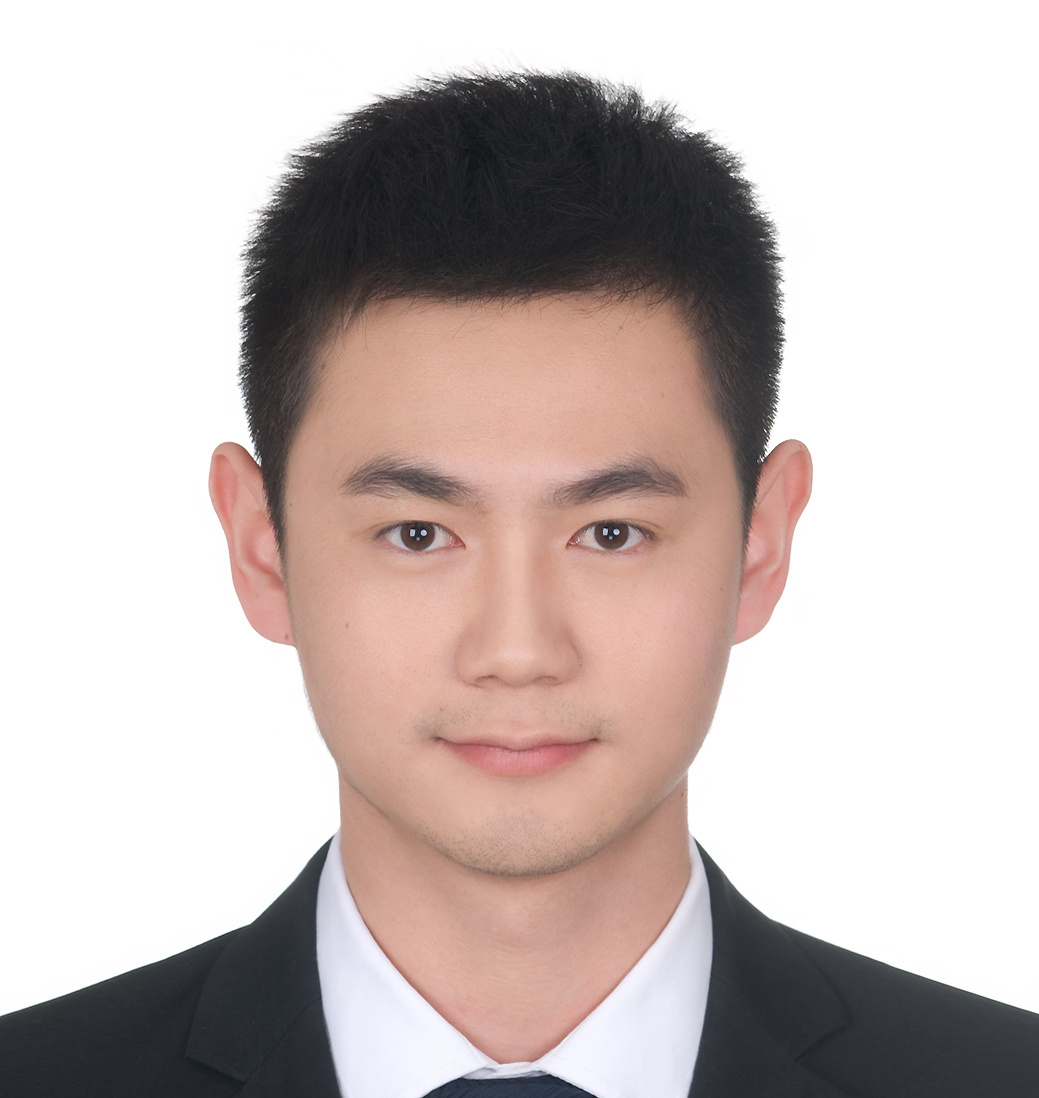}
Wenxiang Jiao is a senior researcher with the Tencent AI Lab, Shenzhen, China. He received his Ph.D. degree from the Chinese University of Hong Kong in 2021, under the supervision of Prof. Irwin King and Prof. Michael R. Lyu. Before that, he received his Bachelor degree and Mphil degree at Nanjing University in 2015 and 2017, respectively. His research interests include conversational emotion recognition, neural machine translation, and multilingual pretraining, and has published papers in top-tier conferences and journals such as ACL, EMNLP, NAACL, AAAI, TASLP, etc. He has won the 1st place at WMT~2022 Large-Scale Machine Translation Evaluation for African Languages (Constrained Track), and the Best Paper Award of Multilingual Representation Learning Workshop in EMNLP 2022.
\endbio

\bio{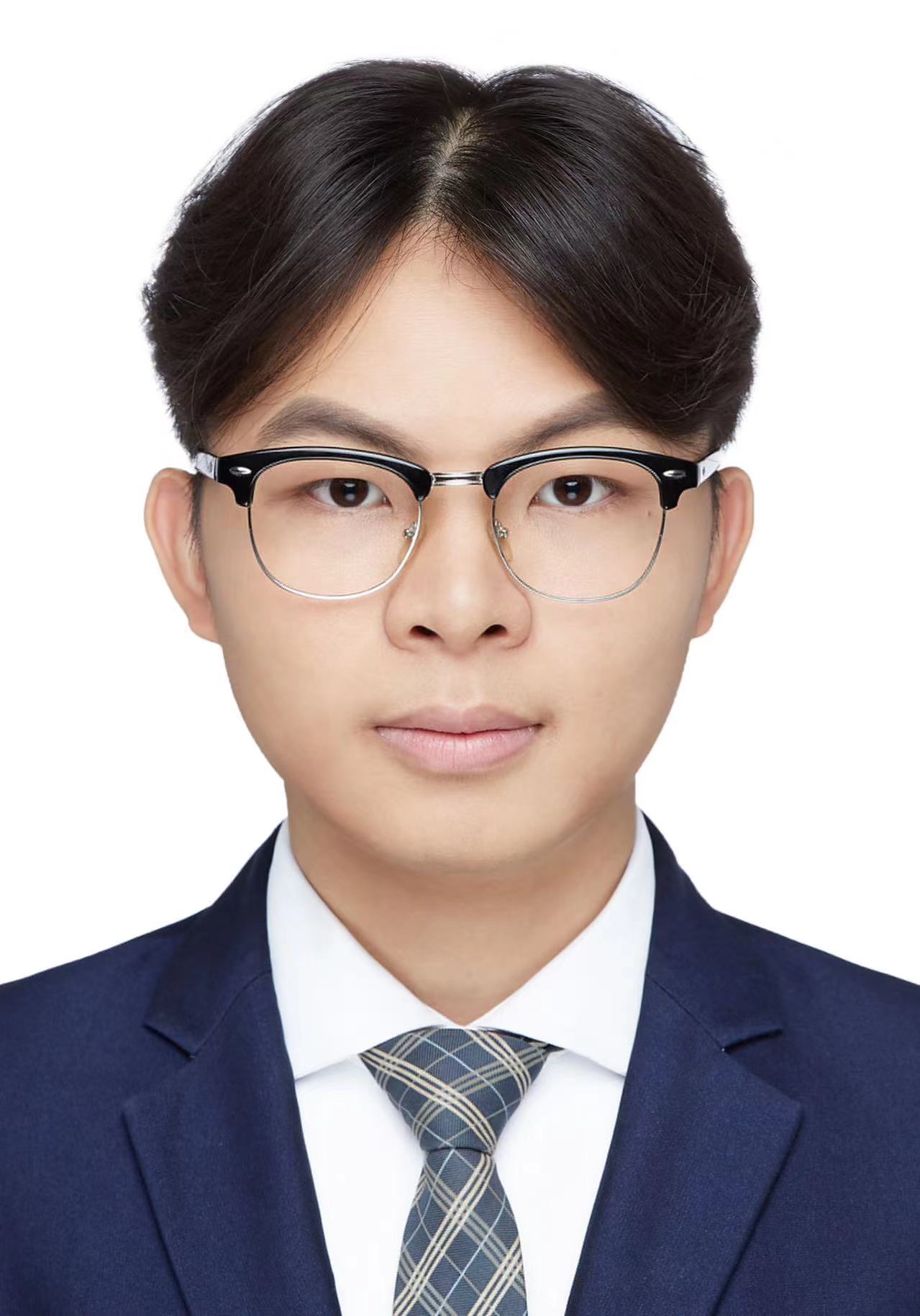}
Jen-tse Huang is a Ph. D. candidate at the Department of Computer Science and Engineering, The Chinese University of Hong Kong, supervised by Prof. Michael R. Lyu. He received his B. S. from Peking University in 2019. His research interest mainly lies in the reliability, robustness, and interpretability of natural language processing models and software.
\\
\\
\endbio

\bio{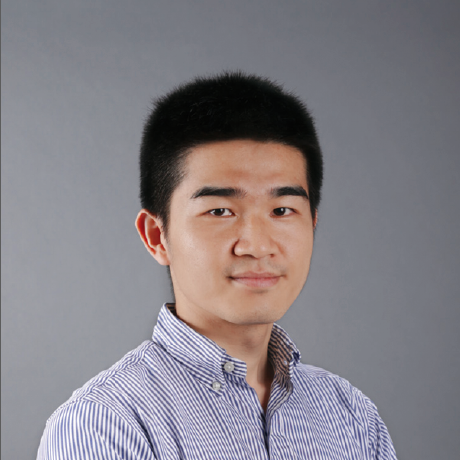}
Zhaopeng Tu is a Principal Researcher with the Tencent AI Lab, Shenzhen, China. He received his Ph.D. degree from Institute of Computing Technology, Chinese Academy of Sciences in 2013. He was a Postdoctoral Researcher at University of California at Davis from 2013 to 2014. He was a researcher at Huawei Noah’s Ark Lab, Hong Kong from 2014 to 2017. He is currently working on neural machine translation and Seq2Seq learning for other NLP tasks, such as dialogue and question answering.
\endbio

\bio{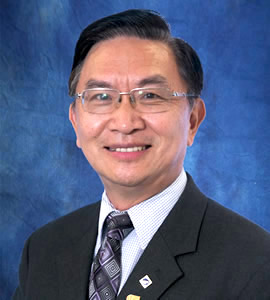}
Michael R. Lyu is currently a Professor at the Department of Computer Science and Engineering in The Chinese University of Hong Kong.
He received his B.S. in Electrical Engineering from National Taiwan University in 1981, his M.S. in Computer Science from University of California, Santa Barbara, in 1985, and his Ph.D. in Computer Science from University of California, Los Angeles in 1988.
His research interests include software engineering, dependable computing, distributed systems, cloud computing, mobile networking, big data, and machine learning.
He was elected to IEEE Fellow (2004), AAAS Fellow (2007), and ACM Fellow (2015) for his contributions to software reliability engineering and software fault tolerance.
\endbio

\end{document}